\documentclass[acmtog]{acmart}

\setcopyright{none}
\renewcommand\footnotetextcopyrightpermission[1]{}

\AtBeginDocument{%
  }

\acmJournal{TOG}
\acmYear{2025}

\usepackage{my}
\usepackage{tikz}
\usepackage{xspace}
\usepackage{overpic}
\usepackage{bbding}
\usepackage{caption} 
\usepackage{float}
\usepackage{subcaption}

\makeatletter
\@namedef{ver@everyshi.sty}{}
\makeatother


\begin{document}

\title{Motion2Motion: Cross-topology Motion Transfer with Sparse Correspondence}

\author{Ling-Hao Chen}
\authornote{Work done during Ling-Hao Chen's internship at IDEA Research. \url{evan@lhchen.top}.}
\orcid{0000-0002-2528-6178}
\email{evan@lhchen.top}
\affiliation{%
  \institution{Tsinghua University, International Digital Economy Academy}
  \country{China}
}

\author{Yuhong Zhang}
\orcid{0000-0001-6180-4457}
\affiliation{%
  \institution{Tsinghua University}
  \country{China}
}

\author{Zixin Yin}
\orcid{0000-0003-0443-7915}
\affiliation{%
  \institution{The Hong Kong University of Science and Technology}
  \city{Hong Kong}
  \country{China}
}

\author{Zhiyang Dou}
\orcid{0000-0003-0186-8269}
\affiliation{%
  \institution{The University of Hong Kong}
  \city{Hong Kong}
  \country{China}
}

\author{Xin Chen}
\orcid{0000-0002-9347-1367}
\affiliation{%
  \institution{ByteDance}
  \city{San Jose}
  \country{United States of America}
}

\author{Jingbo Wang}
\orcid{0009-0005-0740-8548}
\affiliation{%
  \institution{Shanghai Artificial Intelligence Laboratory}
  \city{Shanghai}
  \country{China}
}

\author{Taku Komura}
\orcid{0000-0002-2729-5860}
\affiliation{%
  \institution{The University of Hong Kong}
  \city{Hong Kong}
  \country{China}
}

\author{Lei Zhang}
\orcid{0000-0001-6926-0538}
\affiliation{%
  \institution{International Digital Economy Academy}
  \country{China}
}


\newcommand{\methodname}{\textsc{Motion2Motion}\xspace}
\newcommand{\camera}[1]{{\textcolor{black}{{#1}}}}

\authorsaddresses{}

\renewcommand{\shortauthors}{Ling-Hao Chen, et al.}

\begin{abstract}
  This work studies the challenge of \camera{transfer} animations between characters whose skeletal topologies differ substantially. While many techniques have advanced retargeting techniques in decades, \camera{transfer} motions across diverse topologies remains less-explored. The primary obstacle lies in the inherent topological inconsistency between source and target skeletons, which restricts the establishment of straightforward one-to-one bone correspondences. Besides, the current lack of large-scale paired motion datasets spanning different topological structures severely constrains the development of data-driven approaches. To address these limitations, we introduce \methodname, a novel, training-free framework. Simply yet effectively, \methodname works with only one or a few example motions on the target skeleton, by accessing a sparse set of bone correspondences between the source and target skeletons. Through comprehensive qualitative and quantitative evaluations, we demonstrate that \methodname achieves efficient and reliable performance in both similar-skeleton and cross-species skeleton \camera{transfer} scenarios. The practical utility of our approach is further evidenced by its successful integration in downstream applications and user interfaces, highlighting its potential for industrial applications. Code and data are available at \url{https://lhchen.top/Motion2Motion}.
\end{abstract}

\begin{teaserfigure}
\centering
\includegraphics[width=\textwidth]{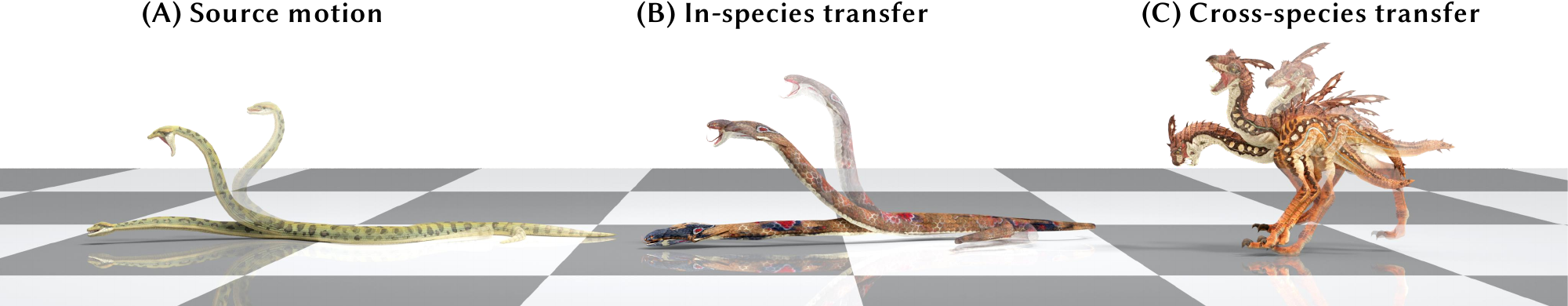}
\caption{We propose \methodname, enabling motion \camera{transfer} across characters with vastly different topologies. From left to right, we show motion \camera{transfer} results across increasingly different target characters, anaconda $\to$ king-kobra (in-species \camera{transfer}) $\to$ T-rex (cross-species \camera{transfer}).}
\label{fig:teaser}
\end{teaserfigure}

\maketitle

\section{Introduction} 
\label{sec:intro}

\camera{Transferring} a motion from one character to another with different topology is a long-standing research problem in computer animation. Motion \camera{transfer} occupies a pivotal position in character animations, especially when mapping motion to unseen characters. To tackle this fundamental problem, a series of methods have been proposed~\citep{aberman2020skeleton,zhang2023skinned} in recent years. One category is skeleton-based motion \camera{transfer} for various shapes of skeletons, the other takes the geometry of two characters into consideration for more fine-grained motion \camera{transfer}.

Despite these progresses, previous methods face significant challenges in real applications, especially \camera{transfer} motions between two characters with different topologies. Regardless of whether skeleton-based or geometric-based methods, most of these methods~\cite{ye2024skinned,zhang2023skinned} rely on a quite amount of data. However, due to the limited accessibility of motion datasets with different characters, these methods often fail at test time. Another trickier issue is that the skeleton topologies and geometries of the target characters are often more complex than the training data. For instance, methods trained on humanoid-like data, such as Mixamo~\citep{mixamo}, struggle to retarget motions to characters with complex dynamics like skirts or hair, let alone \camera{transfer} from bipedal to quadrupeds. Although recent attempts~\citep{walkthedog} tried to retarget animation between humans and quadrupeds, they still need tailored training to accommodate diverse morphologies, making generalization to more complex characters challenging.

There are two key challenges to \camera{transfer} animations from one topology to another. The first challenge is the limited availability of animation data across diverse topologies. In practical applications, only a few motion examples exist for the target skeleton, forcing data-driven methods to rely on large-scale datasets. Unfortunately, even the most readily accessible human motion data, such as SMPL-based motion~\citep{smpl}, suffers from an insufficient volume of high-quality examples. The second challenge involves binding correspondences for all bones between source and target skeletons, especially when the target one has a significantly different topology or number of joints. This makes it difficult to define consistent joint-level mappings, which are critical for motion \camera{transfer}.

To make this problem tractable, we introduce two mild assumptions that align with real-world constraints. First, we assume access to a small number of motion sequences on the target skeleton, a few-shot setting that reflects the scarcity of data in most practical scenarios. If there is no motion example at all, \camera{transfer} becomes ill-posed, as there is no signal to anchor temporal or spatial motion patterns on target skeletons. This minimal data availability provides satisfying contexts to guide meaningful \camera{transfer} while avoiding the need for large-scale annotation. Second, we assume that a sparse joint correspondence between source and target skeletons is available, which can be specified by users or automatic matching (\cref{sec:binding}). This minimal bone mapping offers rough semantic alignment and serves as a useful prior for motion-level matching, a step that is commonly required in both academic and commercial systems (\eg AutoRig~\citep{autorigpro}, and Rokoko~\citep{rokoko}).

Considering these issues, we introduce \methodname, a novel approach for animation \camera{transfer} under sparse joint correspondences (sparse bone binding). Specifically, \methodname models the cross-topology motion \camera{transfer} problem as a conditional patch-based motion matching problem. The bone correspondences specified by users provide coarse motion semantics, serving as the input spatial conditions for \camera{transfer}. 
Departing from traditional neural \camera{transfer} pipelines, \methodname synthesizes motions for the target skeleton by matching motion patches of the bound joints from a few example animations. This design is motivated by a straightforward observation: \textit{motion coordination between bound and unbound joints can often be inferred by observing only a few examples}. For instance, in quadruped locomotion, the movement of the hind legs can often be extrapolated from the motion patterns of the front legs alone by some examples, even in one/few-shot settings. This simple yet effective matching-based \camera{transfer} allows \methodname to outperform current state-of-the-art methods.

A key application of \methodname is its capacity to retarget a SMPL-based motion~\cite{smpl} to those more complex topologies, such as characters featuring dynamic elements like skirts or hair. In addition to this capability, \methodname further enables seamless motion \camera{transfer} across structurally diverse skeletons, even between different species, \textit{e.g.} anaconda \textit{v.s.} raptor (in~\cref{fig:anaconda-king-raptor}). Our framework is also flexible in its choice of motion representation. Specifically, it supports a variety of features for \camera{transfer}, including velocity fields, and the local pose positions, demonstrating its broad generalizability and robustness.

This work paves the way for topology-flexible motion \camera{transfer}, enabling motion adaptation across different character structures in real time. To support practical use, we also develop a Blender add-on (\cref{sec:add-on}), demonstrating the applicability in real-world animation workflows with good interpretability.

\section{Related Work}

\noindent \textbf{\textit{Motion Retargeting.}} 
Motion \camera{transfer} was first introduced by \citet{gleicher1998retargetting}, who approached the problem as a space-time optimization challenge subject to kinematic constraints. This seminal work paved the way for various optimization-based methods that incorporated additional constraints~\citep{bernardin2017normalized,feng2012automating,lee1999hierarchical}. More recently, advances in deep learning have enabled more flexible, data-driven solutions for motion retargeting. For instance, \citet{jang2018variational} proposed an autoencoder-based method for motion generation, \citet{aberman2020skeleton} introduced a Skeleton-Aware Network to address differences in skeletal topology, and \citet{lim2019pmnet} developed a framework that decouples the learning of local poses from global motion. \citet{li2022iterative} presented an iterative retargeting approach that uses motion autoencoders to refine results over multiple steps. Some methods have incorporated character geometry into motion retargeting by preserving contacts or avoiding interpenetration \citep{lyard2008motion,ho2010spatial,jin2018aura,zhang2023skinned,zhang2024semantics}. 
Unlike previous methods that rely on dense correspondences or large-scale data, our approach works under sparse joint mappings and limited target motions. We formulate motion \camera{transfer} as a matching-and-blending process over motion patches, enabling flexible \camera{transfer} across structurally different skeletons. This design allows generalization to complex topologies with minimal samples.

\noindent \textbf{\textit{Generative motion models and motion matching.}} 
Human motion generation~\citep{t2g, temos, motiondiffuse, teach, mdm, humanise, mofusion, physdiff, diffprior, remodiffuse, omnicontrol, humantomato, flowmdm} reaches significant progress in past years, benefiting from the foundational human-centric vision technologies~\citep{smpl,amass}. Researchers have extensively explored the generation of human motion using various modalities—including audio~\cite{t2g,chen2024enabling}, music~\cite{edge,finedance}, and text~\citep{motionlcm,mld,mdm,emdm}—as well as through unconditional approaches~\citep{li2022ganimator,humanmac}. In addition to neural synthesis methods, techniques based on generative motion matching have also shown strong performance on synthesizing high-quality motions~\cite{weiyu23GenMM,buttner2015motion,holden2020learned,bergamin2019drecon}. Motion matching~\camera{\citep{mamm}} itself is a high-level concept in animation creation, primarily employed for character control. However, the application of motion matching and blending to motion \camera{transfer}, especially across different skeletal topologies, remains underexplored from both research and engineering perspectives. The latest progress in generating motions across diverse topologies is represented by AnyTop~\cite{gat2025anytop}. However, AnyTop’s architecture does not include a motion \camera{transfer} module, and its generalization capability is restricted by the relatively limited scale of its training data. In contrast, \methodname takes an early step by \camera{transfer} a source motion to a target skeleton using one or a few sample target motions, marking the first attempt to harness motion matching and blending in cross-topology \camera{transfer}.

\section{Method}

\subsection{Preliminaries}

\noindent \textbf{\textit{Motion representation.}} Let $\mathbf{M} \in \mathbb{R}^{F \times D}$ denote the overall motion representation, where $F$ is the number of frames and $D$ is the dimensionality of the representation. The source motion is represented by $\mathbf{S} \in \mathbb{R}^{F_s \times D_s}$. The target motion set is defined as, $\mathcal{T} = \{ \mathbf{T}^{(i)} \in \mathbb{R}^{F_i \times D_t} \mid i = 1, 2, \dots, N \}$,
where $N$ is the number of target motion sequences and $D_t$ is the dimensionality of each target motion. 
Generally, for the frame $f$ of motion $\mathbf{M}\in \mathbb{R}^{F \times D}$, the pose $\mathbf{M}_{f} \in \mathbb{R}^{D}$ are represented as the 6D joint rotations and the root velocity. Specifically, for a character with $J$ joints, $D \leftarrow 3+6J$.

\noindent \textbf{\textit{Rest pose pre-alignment.}} As the definitions of rest poses in different files are not always the same, we align the protocol of rest poses of source and target motions before \camera{transfer}. The process can be directly borrowed from existing animation tools~\citep{autorigpro}.


\subsection{Motion Patching}

\begin{figure*}[!t]
    \centering
    \begin{overpic}[width=\linewidth]{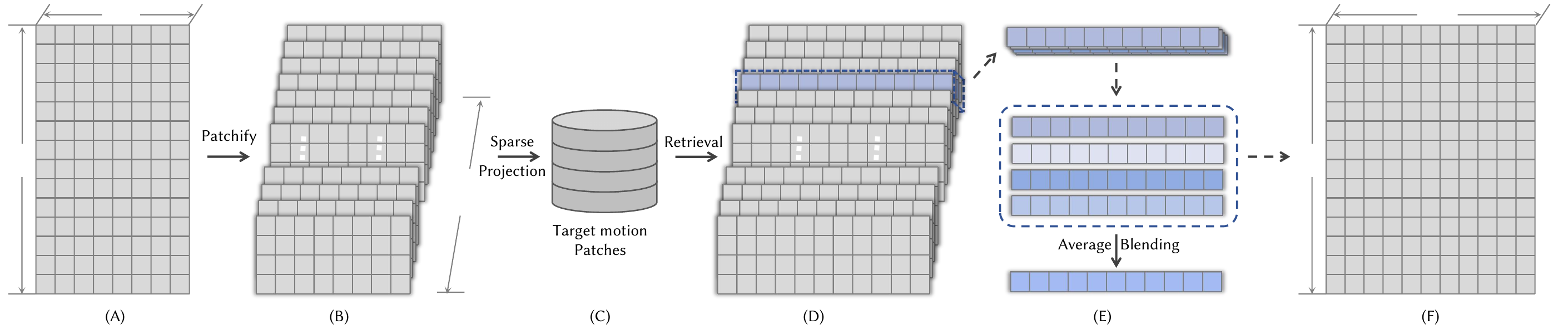}
        \put(6.8, 20.2){\small \textcolor[HTML]{a1a1a1}{\( D_s \)}} 
        \put(0.5, 10.5){\small \textcolor[HTML]{a1a1a1}{\( F_s \)}}  
        \put(29.3, 8.6){\small \textcolor[HTML]{a1a1a1}{\( P \)}}  
        \put(83, 10.5){\small \textcolor[HTML]{a1a1a1}{\( F_s\)}}  
        \put(90.7, 20.2){\small \textcolor[HTML]{a1a1a1}{\( D_t \)}} 
    \end{overpic}
    \caption{
        \textbf{System overview of \methodname.} 
        (A) The source motion sequence \(\mathbf{S} \in \mathbb{R}^{F_s \times D_s}\). 
        (B) The source sequence is divided into overlapping motion patches \(\mathcal{P}^{(s)}\). 
        (C) Each source patch is projected to the target skeleton space via sparse mapping and noise initialization, serving as the query for retrieval. For each source patch, we retrieve target patches (D) from a pre-built motion patch database \(\mathcal{P}^{(t)}\), based on sparse correspondences. 
        (E) The matched target patches are averaged for blending. 
        (F) The  retargeted motion \(\widehat{\mathbf{T}} \in \mathbb{R}^{F_s \times D_t}\) is reconstructed from the blended target patches. (C)-(F) are executed in $L$ times.
    }
    \label{fig:method}
\end{figure*}
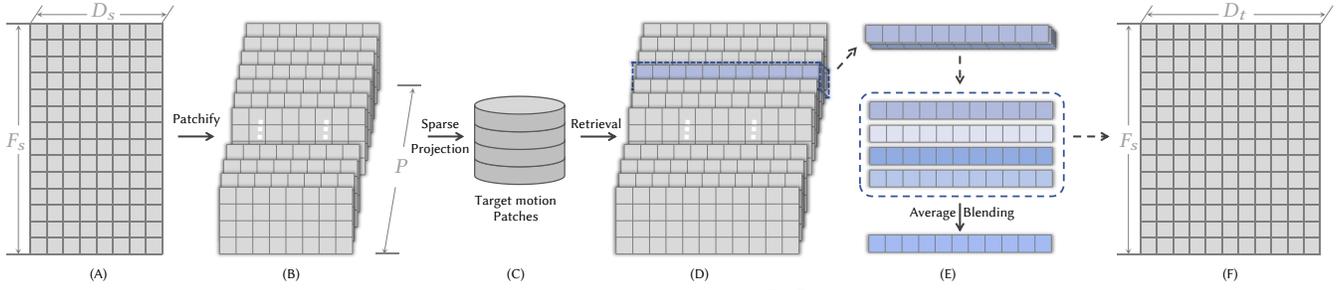

As illustrated in \cref{fig:method}-(A) and (B), prior to \camera{transfer}, we perform motion patching on the source motion along the temporal axis without padding. In contrast to the patching operation in the image domain, we apply a sliding window to patchify the motion, ensuring the preservation of meaningful motion features. This process is grounded in the observation that motion patches capture significant temporal dynamics, and the patch size ($P_S$) is relatively atomic. After the motion patching operation, we obtain $P$ patches of the motion, where $P = \left\lfloor \frac{F_s - \text{patch size} + 1}{\text{step size}} \right\rfloor$. Similarly, we apply the same operation to the target motion set using the same patch and step size, resulting in target motion patches in \cref{fig:method}-(C). The source and target motion patches are denoted as set $\mathcal{P}^{(s)}$ and set $\mathcal{P}^{(t)}$, respectively.

\subsection{Spatial Correspondence between Skeletons}

Let $\mathcal{M} = \{ (t, s) \mid t \in \mathcal{J}_t,\; s \in \mathcal{J}_s \}$ denote the set of sparse keypoint correspondences between the target and source skeletons, where $\mathcal{J}_t$ and $\mathcal{J}_s$ represent the index sets of keypoints in the target and source skeletons, respectively. We denote the number of correspondences by $K = |\mathcal{M}|$.
A correspondence $(t, s) \in \mathcal{M}$ indicates that the keypoint $t$ in the target skeleton is semantically aligned with the keypoint $s$ in the source skeleton. The correspondence of bone binding can be specified by the user or be bound in an automatic way via fuzzy subgraph matching (algorithm detail in \cref{sec:binding}). 

To specify the indexes of each keypoint within the motion representation vectors, we introduce two index functions $\mathcal{I}_1(\cdot), \ \mathcal{I}_2(\cdot): \mathbb{N} \to \mathbb{N}$,
such that for any keypoint $t$ (or $s$), $\mathcal{I}_1(t)$ and $\mathcal{I}_2(s)$ \camera{denote the start and end indices of the channels in the corresponding bones} (\textit{e.g.}, $\mathbf{T}\in\mathbb{R}^{F\times D_t}$ for the target or $\mathbf{S}\in\mathbb{R}^{F\times D_s}$ for the source).

Based on these definitions, we construct a correspondence matrix $\mathbf{C} \in \mathbb{R}^{D_t \times D_s}$ whose block components are defined as follows:
\begin{equation}
\mathbf{C}\Big[\mathcal{I}_1(t):\mathcal{I}_2(t),\; \mathcal{I}_1(s):\mathcal{I}_2(s)\Big] =
\begin{cases}
\ \mathbf{I} \ ,& \text{if } (t, s) \in \mathcal{M},\\
\mathbf{O} \ ,& \text{otherwise},
\end{cases}
\end{equation}
where $\mathbf{I}$ denotes an identity matrix of appropriate dimensions and $\mathbf{O}$ denotes a zero matrix of corresponding size.
This formulation ensures that, in motion matching, only the segments of the motion representations corresponding to the predefined correspondences are aligned, thereby effectively capturing the core kinematic characteristics of the motion while discarding redundant information.

We further introduce a mask vector to identify target skeletal components that lack any correspondence in $\mathcal{M}$. Specifically, the mask is computed as $\mathbf{m}[i] = \sum_{j=1}^{D_s} \mathbf{C}[i, j]$, which effectively indicates, for each target dimension $i$, whether a valid mapping from the source exists. Note that the vector $\mathbf{m}$ can be sparse, that is a small number of joint correspondences.

\subsection{Iterative Matching-based \camera{Transfer}}
\label{sec:retarget_alg}

With the sparse correspondence established above, we aim to \camera{transfer} motion information between structurally different skeletons. Motivated by motion matching, we formulate the \camera{transfer} process as a retrieval-and-blending procedure over motion patches. This allows us to explore the application of motion matching under cross-topology settings with sparse features, where the source and target skeletons differ significantly in structure and representation.

\noindent \textbf{\textit{Motion projection and initialization.}} Using this mask, the mapping from the source motion  to the target skeleton is defined as
\begin{equation}
    \mathbf{P}^{s \to t} = \mathbf{S}\mathbf{C}^\top + \bigl(\mathbf{1} - \mathbf{m}\bigr) \odot \mathbf{N},
    \label{eq:proj}
\end{equation}
where $\mathbf{N} \sim \mathcal{N}(\mathbf{O}, \mathbf{I}) \in \mathbb{R}^{F \times D_t}$ denotes a noise term drawn from a standard normal distribution. The first term, $\mathbf{S}\mathbf{C}^\top$, represents the motion mapping from the source skeleton to the target skeleton based on predefined keypoint correspondences. The second term, $\bigl(\mathbf{1} - \mathbf{m}\bigr) \odot \mathbf{N}$, is initialized by a noise for target dimensions without a corresponding source mapping. The \camera{transfer} process can be viewed as the transformation of these noisy components into the desired locomotion, conditioned on the first term. After the procedure, the noisy parts will be replaced by the retargeted locomotion. This formulation ensures that for target dimensions without a corresponding source mapping, the motion is augmented by controlled stochastic variation. For convenience, we rename $\mathbf{P}^{s \to t}$ as $\widehat{\mathbf{T}}$ in following sections.

\noindent \textbf{\textit{Masked motion matching.}} After obtaining the projected motion $\widehat{\mathbf{T}}$ in \cref{eq:proj}, we formulate the synthesis process of the noisy part as a generative process conditioning on $\mathbf{S}\mathbf{C}^\top$. We also patchify the the projected motion $\widehat{\mathbf{T}}$ as $\mathcal{P}^{(\widehat{t})}$. Motivated by generative motion matching~\cite{weiyu23GenMM,buttner2015motion}, the generation process of target motion patches in \methodname is masked motion patch retrieval. Specifically, for each $\mathbf{P}^{(\widehat{t})} \in \mathcal{P}^{(\widehat{t})}$,
\begin{equation}
    \begin{aligned}
        \mathbf{P}^{match} \leftarrow \mathop{\arg\min}\limits_{\mathbf{P} \in \mathcal{P}^{(t)}} &\;\; \alpha \mathcal{L} \left( \mathbf{m}\odot \mathbf{P}, \mathbf{m}\odot \mathbf{P}^{(\widehat{t})} \right) \\ 
        &+ (1-\alpha)\mathcal{L} \left((\mathbf{1}- \mathbf{m})\odot \mathbf{P}, (\mathbf{1}- \mathbf{m})\odot \mathbf{P}^{(\widehat{t})} \right),
    \end{aligned}
    \label{eq:match}
\end{equation}

\noindent where $\mathcal{L}(\cdot, \cdot)$ denotes the mean squared error (MSE) loss, and $\alpha \in [0, 1]$ balances the contributions from mapped and unmapped parts in $\widehat{\mathbf{T}}$. A larger $\alpha$ in \cref{eq:match} places more emphasis on the sparse correspondences, enforcing stronger semantic alignment between source and target motions. In contrast, the second term leverages noise in unmapped dimensions, introducing diversity and flexibility into the generated motion.

{
\setlength{\textfloatsep}{0em} 
\begin{algorithm}[t]
\SetAlgoLined
\caption{Iterative Matching-based Motion Retargeting}
\label{alg:transfer}
\KwIn{Source motion $\mathbf{S}$, target patch set $\mathcal{P}^{(t)}$, weight $\alpha$, correspondence matrix $\mathbf{C}$, mask $\mathbf{m}$, iterations $L$.}
\KwOut{Retargeted target motion $\widehat{\mathbf{T}}$.}


Project source motion: $\mathbf{P}^{s \to t} \leftarrow \mathbf{S} \mathbf{C}^\top + (\mathbf{1} - \mathbf{m}) \odot \mathbf{N}$.\\
Initialize $\widehat{\mathbf{T}} \leftarrow \mathbf{P}^{s \to t}$.\\
\For{$\ell \leftarrow 1$ \KwTo $L$}{
    Initialize matched target set $\mathcal{M}=\{\}$.\\
    Pathify $\widehat{\mathbf{T}}$ into $\mathcal{P}^{(\widehat{t})}$.\\
    \ForEach{$\mathbf{P}^{(\widehat{t})} \in \mathcal{P}^{(\widehat{t})}$}{
        Retrieve $\mathbf{P}^{^{match}}$ from $\mathcal{P}^{(t)}$ with mask $\mathbf{m}$ by \cref{eq:match}.\\
        Append $\mathbf{P}^{match}$ into $\mathcal{M}$.\\
    }
    Blend patches in $\mathcal{M}$ as $\widehat{\mathbf{T}}$.\\
}
\Return{$\widehat{\mathbf{T}}$}
\end{algorithm}
}

\noindent \textbf{\textit{Motion blending.}} For each source motion patch $\mathbf{P}^{(s)}\in\mathcal{P}^{(s)}$, we match a similar target patch $\mathbf{P}^{^{match}}\in\mathcal{P}^{(t)}$ according to \cref{eq:match}, resulting a matching set $\mathcal{M}$. We blend all retrieved target motion patches of $\mathcal{M}$ in average, as illustrated in \cref{fig:method}-(E). 

However, directly blending the retrieved patches may compromise motion naturalness or distort key actions, especially in the regions corresponding to the sparse correspondences. To address this, we repeat the matching-and-blending process $L$ times, progressively refining the result to ensure temporal coherence across the entire sequence. The whole algorithm is in \cref{alg:transfer}.

\begin{table*}[!t]
    \centering
    \resizebox{\textwidth}{!}{
    \begin{tabular}{c|c|ccccc|ccccc}
    \toprule
         \multirow{2}{*}{Method} & \multirow{2}{*}{training} & \multicolumn{5}{c}{Similar Skeleton} & \multicolumn{5}{|c}{Cross-species Skeleton} \\ \cline{3-12}
         & & FID$\downarrow$ & freq. align$(\%)$$\uparrow$ & contact con.$(\%)$$\uparrow$ & diversity$\uparrow$ & FPS$\uparrow$ & FID$\downarrow$ & freq. align$(\%)$$\uparrow$ & contact con.$(\%)$$\uparrow$ & diversity$\uparrow$ & FPS$\uparrow$ \\ 
    \midrule
         \citet{walkthedog} & \CheckmarkBold & 0.507 & 72.0 & 70.5 & 0.02 & \textbf{833} & 2.250 & 66.9 & 62.4 & 0.03 & \textbf{729} \\
         \citet{pose2motion} & \CheckmarkBold & 0.389 & 80.1 & 78.6 & 0.33 & 234 & 1.68 & 72.4 & 64.5 & 0.30 & 215 \\
         \methodname &\XSolidBrush & \textbf{0.033} & \textbf{96.2} & \textbf{93.5} & \textbf{3.20} & 778 & \textbf{0.492} & \textbf{90.3} & \textbf{79.7} & \textbf{1.90} & 752 \\
    \bottomrule
    \end{tabular}
    }
    \caption{\textbf{Main evaluation results for motion transfer.} Our method (\methodname), in both similar skeleton and cross-species settings, achieves the best performance on synthesis quality, temporal coherence, and diversity. Different from baselines, our method runs without GPUs and deep model training. }
    \label{tab:mainres}
\end{table*}

\section{Experiments}

\subsection{Implementation Details}

\noindent \textbf{\textit{Data.}} In this section, we collect a set of animal motions and human motions to present motion \camera{transfer} across skeletons. The animal animation dataset is the Truebones-Zoo dataset~\citep{truebones_zoo_2022}. The number of joints ranges from 9 to 143. The human animation data is borrowed from LAFAN~\citep{lafan}. For qualitative experiments, we choose a subset of Truebones-Zoo as the benchmark for comparison. 

\noindent \textbf{\textit{Implementation.}} The patch size of the motion is set as 11 frames, $\alpha$ is set as 0.85 by default. We follow \citet{li2022ganimator}, generating motion in an inverted pyramidal way. All of the baselines mentioned in the paper is run on $1\times$ NVIDIA RTX-3090-24GB GPU, while our method is run on $1\times$ MacBook-M1 chip without a GPU.

\begin{figure*}[!t]
    \centering
    \begin{overpic}[width=\linewidth]{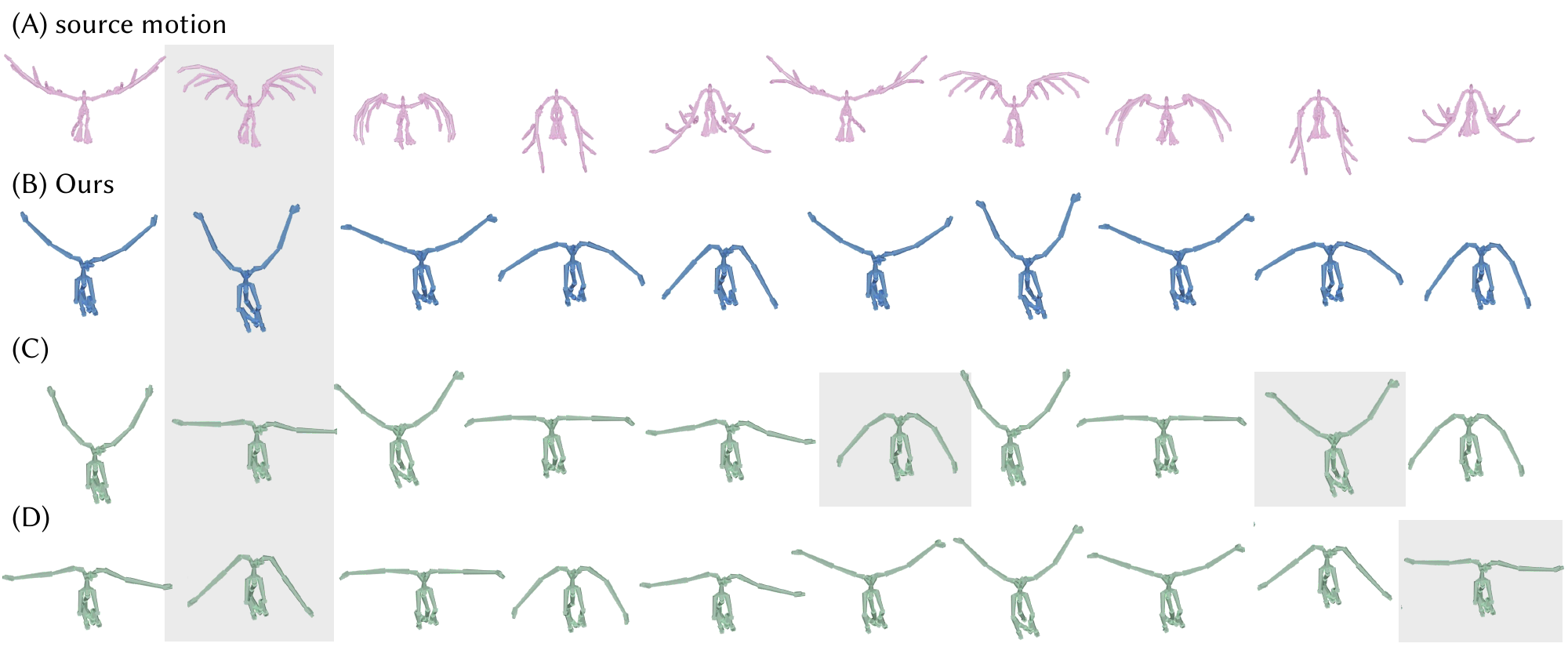}
        \put(1.6, 18.5){\textsf{ \ \ \ \ \citet{walkthedog}}}
        \put(1.6, 7.7){\textsf{ \ \ \ \ \citet{pose2motion}}}
    \end{overpic}
    \caption{\textbf{Quantitative comparison with baselines.} 
    Each animation sequence is listed as frames from left to right. (Highlighted gray frames as comparison for baselines.)
    (A) Input source motion of a dragon character. 
    (B–D) Retargeted sequences on the bat skeleton produced by (B) our method, (C) \citet{walkthedog}, and (D) \citet{pose2motion}.  
    Compared to baselines, our approach more faithfully preserves the original motion style, coherence, and frequency.}
    \label{fig:baseline}
\end{figure*}

\subsection{Evaluation Protocol}
\label{sec:evlprotocol}

\noindent \textbf{\textit{Metrics.}} We evaluate the \camera{transfer} result from the following aspects. (1) Motion quality is evaluated via the Fr\'echet Inception Distance (FID) between retargeted and real target skeleton motions. The FID metric measures the motion quality and the style consistency with the target motion distribution. We use the kinematic features~\camera{\cite{bailando}} as extracted features when calculating FID. 
(2) The temporal alignment between the source and retargeted motion is evaluated by the cosine similarity of the Power Spectral Density~(PSD) across all joints. We also follow \citet{pose2motion} using contact consistency as the metric to evaluate the temporal coherence, where the contact bones are labeled manually by researchers. 
(3) Transfer diversity is the average joint distance between each two samples among the 5 retargeted results. 
(4) We use inference FPS to evaluate efficiency. In remaining text, we calculate the \textbf{binding rate} as $\frac{2|\mathcal{M}|}{J_S+J_T}\times 100\%$, where $J_S$ and $J_T$ are bone numbers of source and target skeletons. 

\noindent \textbf{\textit{Baselines.}} We introduce two latest baselines for comparison. \textit{Pose-to-Motion}~\citep{pose2motion} \camera{transfers} motion from a motion-rich source to a target with only pose data, using a learned pose prior to generate temporally coherent motions. It performs well with sparse or noisy poses and in cross-skeleton settings. In our comparison, we split the motion data into a poses set in ahead. \textit{WalkTheDog}~\citep{walkthedog} aligns motions across different morphologies by a shared phase manifold. It enables semantic and temporal alignment between structurally different characters (e.g., human to quadruped) and supports tasks like motion \camera{transfer} and stylization.

\noindent \textbf{\textit{Benchmark.}} Inspired by previous in-topological motion \camera{transfer} evaluation, we collect 14 character animations to evaluate the cross-topology motion \camera{transfer} results. The evaluation benchmark comes up with 1,167 frames in total, covering running, walking, jumping, and attacking actions from Truebones-Zoo~\citep{truebones_zoo_2022} and LAFAN~\citep{lafan}. To thoroughly evaluate the algorithm bounds, we categorize the skeleton gap between the source and target characters as similar and cross-species \camera{transfer}. In detail, we categorize the source-target characters into similar skeletons (\textit{e.g.}, biped-to-biped or crawling-to-crawling) and cross-species skeletons (\textit{e.g.}, biped-to-quadruped or crawling-to-biped).

\subsection{Comparison with Baselines}

\noindent \textbf{\textit{Qualitative results.}}
As shown in \cref{fig:baseline}, we present a visual comparison of motion \camera{transfer} results across different methods. The top row (A) illustrates the source motion performed by a dragon (143 joints), showcasing large-amplitude wing flaps and dynamic movement patterns. The following rows show the motion retargeted onto a bat (48 joints) using (B) our method (binding 2 pairs, \camera{left and right UpperArm}), (C) Walk-The-Dog \citep{walkthedog}, and (D) Pose-to-Motion \citep{pose2motion}. 
Our method (B) clearly follows the full range of the source motion, preserving both temporal structure and stylistic characteristics such as up and down wing spread events. In contrast, motions produced by baselines suffer from artifacts and temporal inconsistency. Furthermore, our method shows consistent spatial-temporal transitions across frames, demonstrating superior coherence and fidelity in cross-topology \camera{transfer}.

\noindent \textbf{\textit{Quantitative results.}} \cref{tab:mainres} reports the quantitative comparison between our method and two recent baselines, Walk-The-Dog and Pose-to-Motion. Across both similar skeleton and cross-species settings, our method consistently achieves the best performance in all quality and diversity metrics. Specifically, we obtain the lowest FID, indicating high motion quality and style consistency. Our method also achieves the highest frequency alignment and contact consistency, demonstrating strong temporal coherence with the source motion. In terms of diversity, we outperform baselines by a large margin, suggesting the ability to generate varied retargeted motions. Despite being training-free, our method maintains a high inference FPS, comparable to Walk-The-Dog and significantly faster than Pose-to-Motion. Notably, two baselines are tested on one GPU, while ours is without a GPU. Moreover, our method is a model-free algorithm, without any model training. 

\noindent \textbf{\textit{Analysis.}} The reason \textit{why} two baselines work worse than the proposed method is that the model-based methods are highly data-driven, requiring a large amount of data. This not only results in some artifacts on some OOD scenarios, but also easily leads to over-fitting issues and makes the result less diverse (see \cref{tab:mainres}). Moreover, baselines are highly reliant on the trained data distribution. If the desired motion frequency is unobserved in training, it is hard to retarget motion to a novel frequency provided by the source motion (see freq. align \& contact con. in \cref{tab:mainres}). However, our matching-based method can flexibly compose the motion patches by motion patch blending, simulating as a frequency interpolator.

\subsection{Temporal Matching Visualization}

\begin{figure}[!t]
    \centering
    \includegraphics[width=\linewidth]{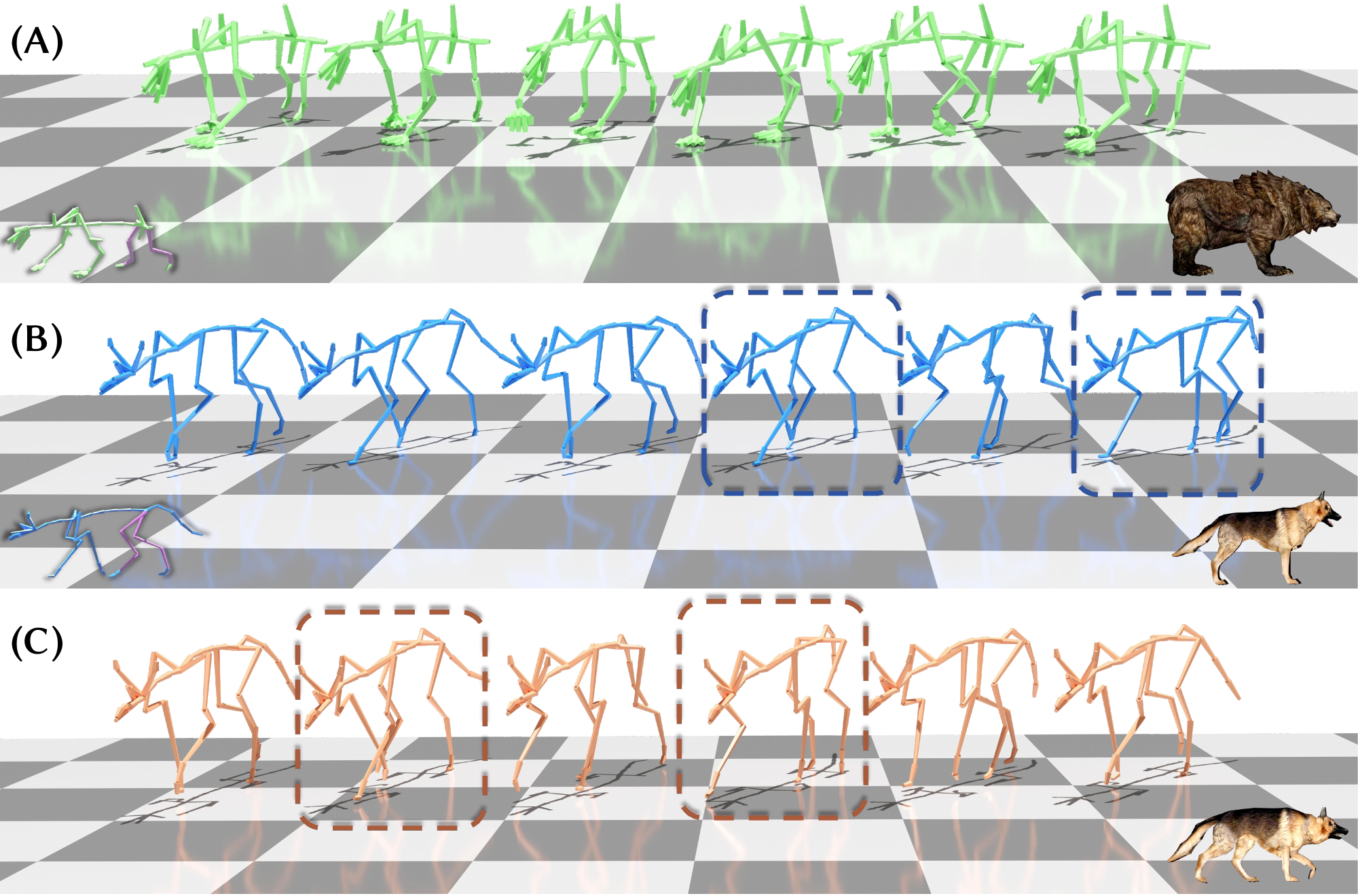}
    \caption{
        \textbf{Motion matching visualization.} 
        A) Source motion on a bear skeleton. (B) Target dog skeleton used for matching. (C) The retargeted motion result on the dog skeleton. The 2nd and 4th frames in (C) correspond to the 4th and 6th frames in (B), respectively, illustrating temporally non-linear yet semantically aligned motion matching.
    }
    \label{fig:bear-dog2}
\end{figure}

\noindent \textbf{\textit{Matching patches visualization.}} 
\cref{fig:bear-dog2} provides a qualitative visualization of our motion \camera{transfer} process. The source and target skeletons correspond to a bear (76 joints) and a dog (55 joints), respectively. As shown on the left side of panels (A) and (B), the correspondences between the two skeletons are limited to 6 hind leg joints. Notably, the second and fourth frames in panel (C) correspond to the fourth and sixth frames in panel (B), demonstrating effective temporal alignment. Moreover, the motion of the dog's front legs is conditionally inferred from its hind leg motion, and the dynamic motion of the dog's long tail is not directly provided by the source. This flexible matching is achieved by leveraging sparse keypoint bindings and masked patch comparisons, enabling robust motion \camera{transfer} across diverse topology skeletons.

\begin{figure}[!t]
    \centering
    \includegraphics[width=\linewidth]{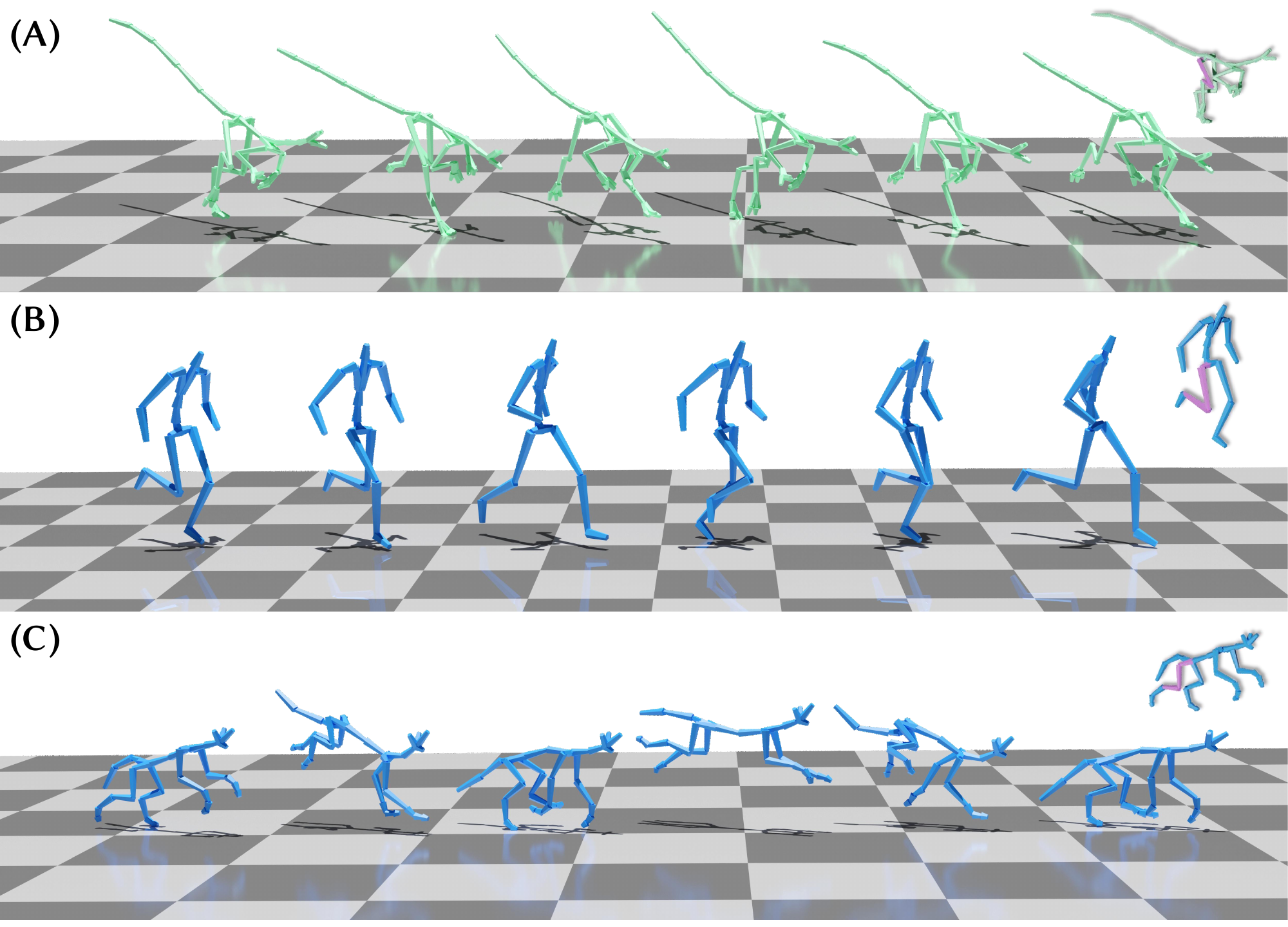}
    \caption{
        \textbf{Motion temporal coherence visualization.} 
        (A) Source running motion from a Raptor character (36 joints). (B) Retargeted motion to a human character (22 joints). (C) Retargeted motion to a fox character (40 joints). The retargeted motions share the same temporal coherence and movement periodicity. \camera{Binding bones are in purple (right side)}.
    }
    \label{fig:fre1-12N}
\end{figure}

\begin{figure}[!t]
    \centering
    \includegraphics[width=\linewidth]{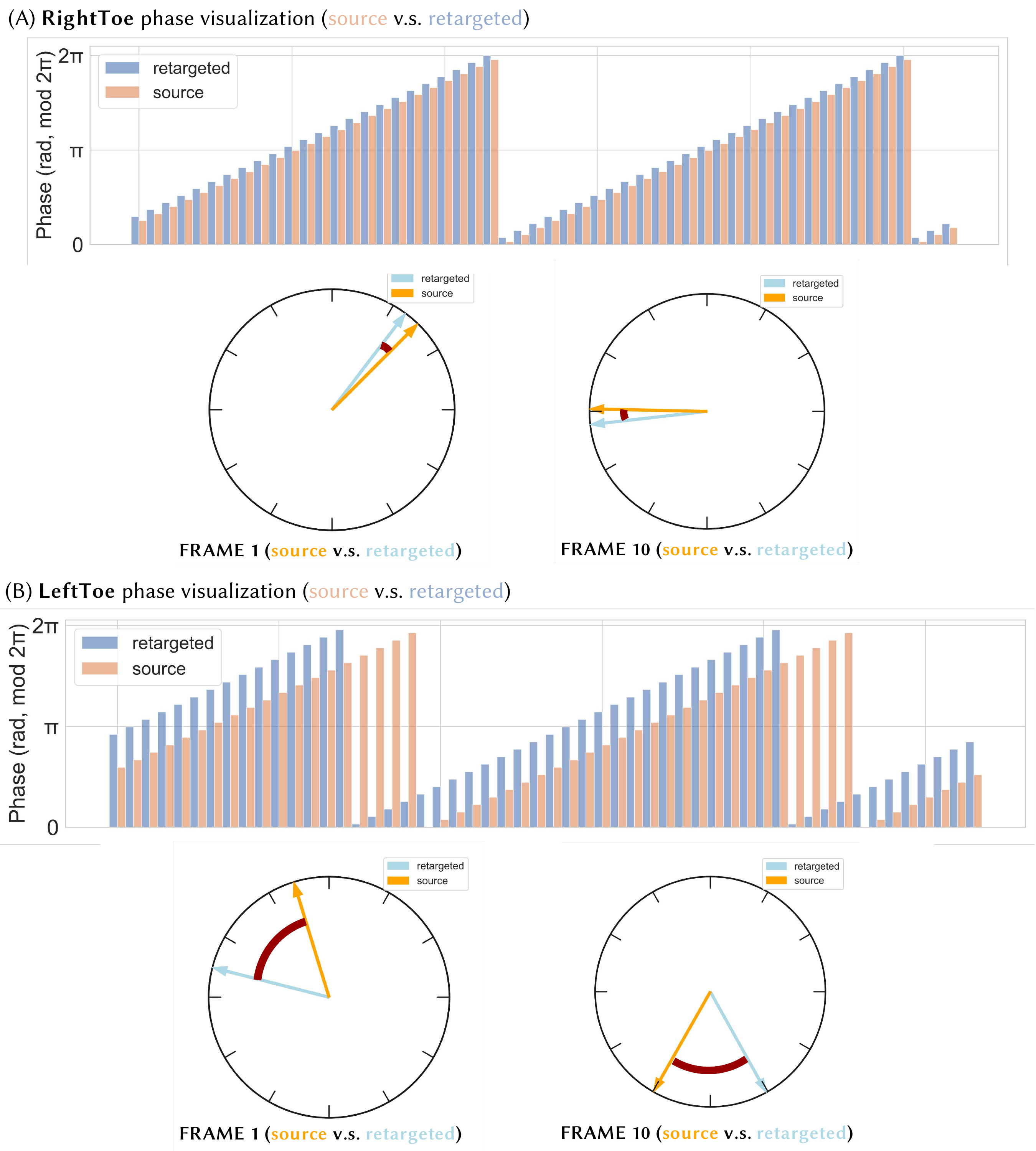}
    \caption{
        \textbf{Phase visualization of the motion.} (A) and (B) present the phases of \texttt{RightToe} and \texttt{LeftToe}. The bar figure is the phase variation curve, and the clock figure is the phase visualization at the 1-st and 10-th frames. The blue and orange colors denoted retargeted and source motion, respectively. Note that there is a consistent phase bias between the source and target. 
    }
    \label{fig:phase}
\end{figure}

\noindent \textbf{\textit{Phase manifold coherence.}}
In addition to maintaining temporal coherence, our method ensures that the retargeted motions lie on a smooth and consistent phase manifold. To present the robustness of the algorithm to human characters (from LAFAN dataset~\citep{lafan}) outside the animal motion dataset. We randomly clip 5 motion sequences from the LAFAN dataset, each of which is no more than 80 frames (data with 60 FPS). 
Previous research~\citet{} shows the periodicity of the motion, especially in the phase manifold. To verify the motion coherence of the source motion and the retargeted one, we visualize the periodic phase of the joint positions. In \cref{fig:phase}, we visualize the phase of the dominant frequency component (\textit{a.k.a.} the frequency component with the largest amplitude in the FFT). The constant phase bias shown in \cref{fig:phase} means the coherent motion frequency. 
As visualized in \cref{fig:fre1-12N}, the phase progression in the retargeted motions (B and C) follows the same \textit{periodicity} as the source Raptor motion (A), despite substantial differences in skeleton topology and motion dynamics. This coherence across species and motion styles highlights the robustness of our framework in preserving underlying phase structures during motion \camera{transfer}.

\subsection{Cross- Skeleton and Species \camera{Transfer}}

\begin{figure}
    \centering
    \includegraphics[width=\linewidth]{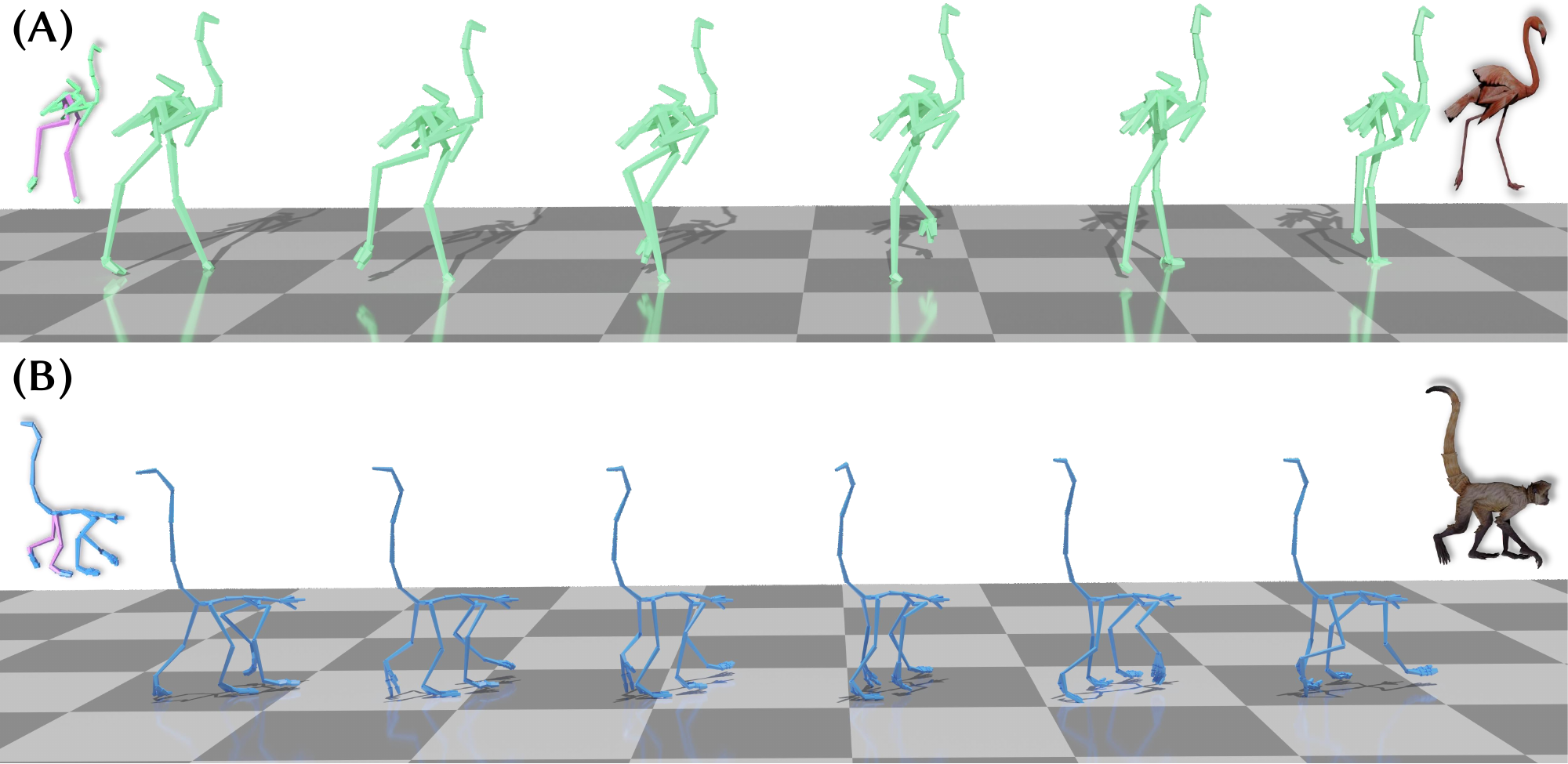}
    \caption{\textbf{Cross skeletal organisms \camera{transfer} (biped $\to$ quadruped).} A walking motion is \camera{transfer}red from a flamingo (41 joints) to a monkey (76 joints) by binding only 6 corresponding hind limb bones. The hind limb motions are kept synchronous, while the monkey's forelimb and tail movements are inferred based on the retargeted hind limb dynamics.}
    \label{fig:walk-flamingo-monkey}
\end{figure}

Cross-skeletal motion \camera{transfer} between species with significantly different anatomical structures is a challenging problem. Our approach resolve this issue with very sparse correspondences. 


\noindent \textbf{\textit{Biped $\to$ quadruped.}}
As illustrated in \cref{fig:walk-flamingo-monkey}, we \camera{transfer} a walking motion from a flamingo, a bipedal species with 41 joints, to a quadrupedal monkey skeleton comprising 76 joints. In this case, we establish sparse keypoint correspondences by binding only 6 hind limb bones between the source and target skeletons. These sparse bindings serve as anchors for \camera{transfer} motion cues. In our framework, the synchronous motion of the hind limbs is maintained between the two skeletons; that is, the source motion directly drives the retargeted hind limb dynamics in the monkey. Meanwhile, the movements of the monkey's forelimbs, head, fingers, and tail are inferred from these sparse hind limb dynamics, ensuring a natural and coherent overall gait despite the structural disparity.

\begin{figure}[!t]
    \centering
    \includegraphics[width=\linewidth]{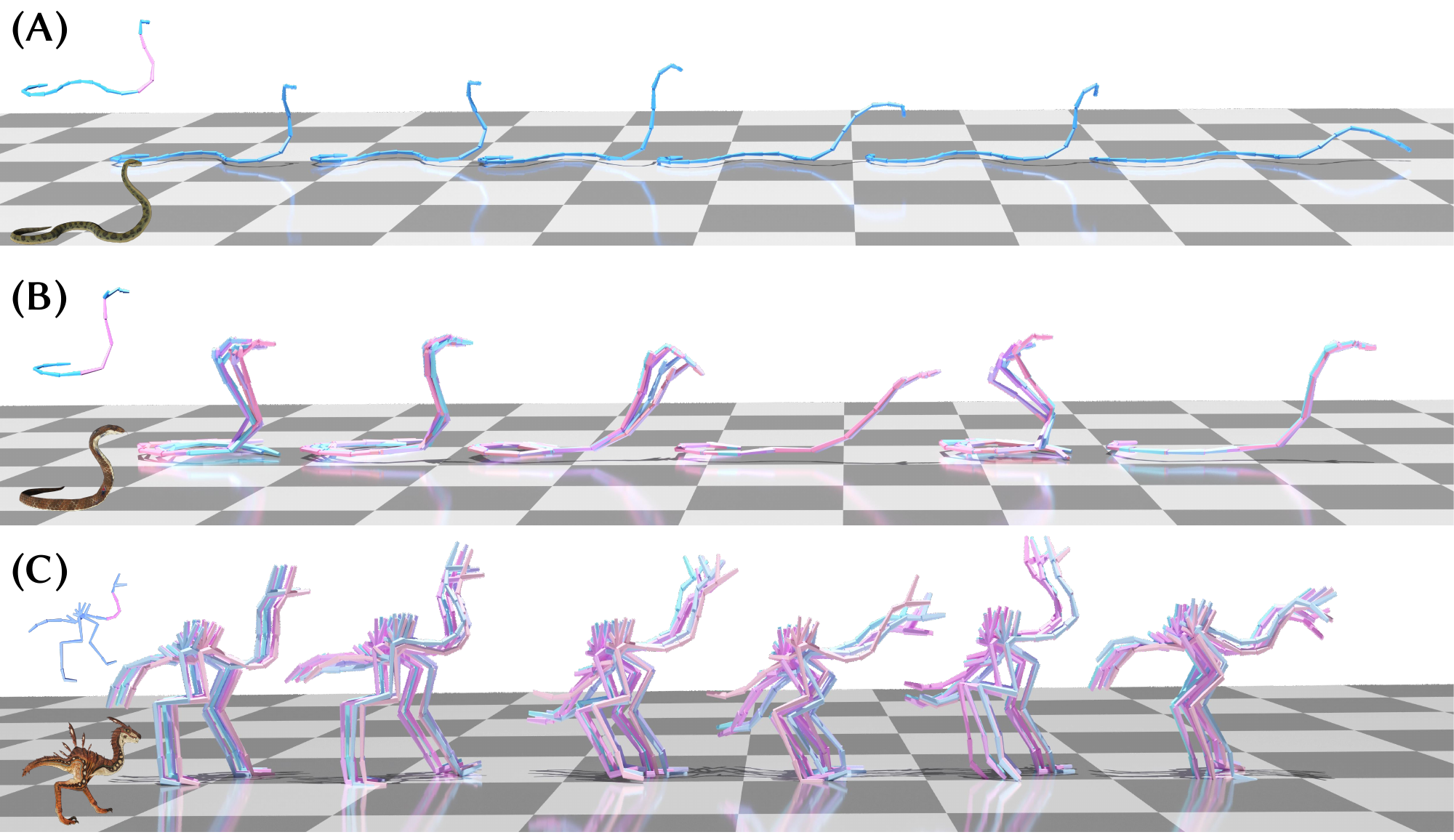}
    \caption{
        \textbf{Cross skeletal organisms \camera{transfer} (limbless $\to$ biped).} The diverse retargeted results of the same skeleton are shown in different colors.         
        (A) Source motion of an anaconda. (B) Retargeted motion on the limbless skeleton (king-cobra). (C) Retargeted motion on the biped skeleton (raptor). The frames in (B) and (C) correspond to specific moments from (A), demonstrating the transformation from a limbless to a bipedal structure with semantically and temporally aligned motion.
    }
    \label{fig:anaconda-king-raptor}
\end{figure}

\noindent \textbf{\textit{Limbless $\to$ biped.}}
We first retarget an attack motion from anaconda (27 joints) to the kingcobra (19 joints) in \cref{fig:anaconda-king-raptor}-(A)/(B), where the retargeted kingcobra motion keeps its original huddling style. As shown in \cref{fig:anaconda-king-raptor}-(A)/(C), we retarget a motion from a limbless species, an anaconda, to a bipedal raptor skeleton with 36 joints. In this case, we establish sparse keypoint correspondences by binding only 4 key vertebral points between source and target skeletons. These sparse bindings are crucial in guiding the \camera{transfer} process, ensuring motion alignment across skeletons. Moreover, as discussed in \cref{sec:retarget_alg}, the \camera{transfer} diversity is controlled by the noise term. As shown in \cref{fig:anaconda-king-raptor}-(B)/(C), different colors for retargeted skeletons represent diverse results driven by the same source motion. 

\noindent \textbf{\textit{Diversity of retargeted motions.}} In \cref{eq:match}, $\alpha \in [0, 1]$ controls the weights of joint correspondence matching and random noise matching. The larger $\alpha$ means more accurate local motion matching on specified joints, that is, less diversity. \cref{fig:anaconda-king-raptor} shows diverse results of the cross-species motion \camera{transfer}. As can be seen in \cref{tab:mainres}, our method enjoys the best diversity over baselines, qualitatively verifying the superiority of our algorithm.

\begin{table}[!t]
    \centering
    \resizebox{0.9\linewidth}{!}{
    \begin{tabular}{c|ccccc}
    \toprule
     Settings & FID$\downarrow$ & freq. align$(\%)$$\uparrow$ & contact con.$(\%)$$\uparrow$ & diversity$\uparrow$\\ 
     \midrule
     Ours ($\times 3$ samples) & 0.230 & 77.5 & 89.7  & 4.50 \\
     Ours ($\times 2$ samples) & 0.237 & 75.9 & 88.0  & 3.16 \\
     \hline
     Ours  & 0.263 & 73.3 & 86.6 & 2.55 \\
    \bottomrule
    \end{tabular}
    }
    \caption{\textbf{``Test time scaling'' property.} The comparison with different number of target samples.}
    \label{tab:scaling}
\end{table}

\subsection{``Test-time Scaling'' Property}
\cref{tab:scaling} reports the performance of our method \camera{when varying the number of target samples in the dataset during inference}. 
We observe that increasing the number of samples improves the overall performance across all metrics. 
Specifically, generating 3 samples significantly reduces the FID score and enhances both frequency alignment and contact consistency, indicating better quality and realism. 
Furthermore, the diversity score increases notably, showing that more samples lead to more diverse motion generations. 
Although the larger number of target examples benefits the \camera{transfer} quality, the more \#frames of the target skeletons help the overall quality.  
This demonstrates that our method benefits from test-time (inference) sampling without training, showcasing strong scalability.


\subsection{User Study}

We evaluate the motion \camera{transfer} result across the following aspects through a user study. Users are required to evaluate the \camera{transfer} result through (1) the quality of retargeted motion. (2) action alignment with the source motion. In this study, \camera{50} users rate 10 source retargeted motion pairs with score 1-5. Baselines are introduced in \cref{sec:evlprotocol}. As shown in \cref{tab:user}, our method leads baseline with a significant margin on both quality and action alignment.

\begin{figure*}[!t]
    \centering
    \includegraphics[width=\linewidth]{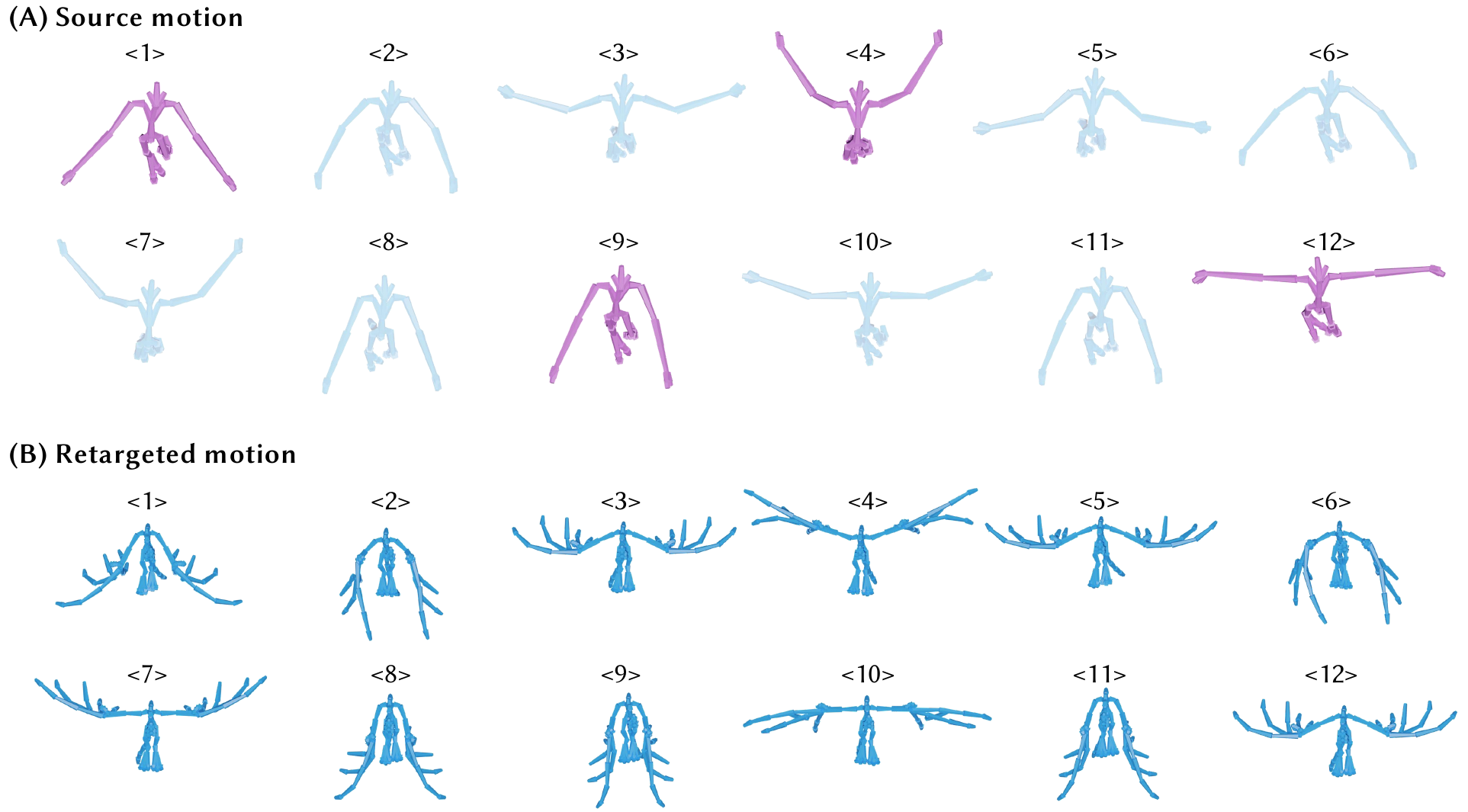}
    \caption{
        \textbf{Key-frame Cross-topology Retargeting.} The frame number index is denoted as ``<\texttt{X}>''. (A) The source motion of a bat. The purple frames are given key frames. The transparent blue frames are ground-truth frames of the source motion. (B) Retargeted motion of a dragon character. 
    }
    \label{fig:key-frame}
\end{figure*}

\begin{table}[!t]
    \centering
    \setlength{\tabcolsep}{9pt}{
    \resizebox{0.85\linewidth}{!}{
    \begin{tabular}{c|ccc}
    \toprule
      & \citet{walkthedog} & \citet{pose2motion} & Ours\\ 
     \midrule
     quality  & \camera{3.55$^{\pm .22}$} & \camera{2.95$^{\pm .10}$} & \camera{\textbf{4.36}$^{\pm .18}$}\\
     alignment  & \camera{3.30$^{\pm .35}$} & \camera{2.84$^{\pm .19}$} & \camera{\textbf{4.60}$^{\pm .11}$} \\
    \bottomrule
    \end{tabular}
    }}
    \caption{\textbf{User study results.} Our method leads baselines over a significant margin on both motion quality and alignment.}
    \label{tab:user}
\end{table}

\begin{table}[!t]
    \centering
    \resizebox{0.9\linewidth}{!}{\setlength{\tabcolsep}{6pt}{
    \begin{tabular}{c|ccccc}
    \toprule
     Settings & FID$\downarrow$ & freq. align$(\%)$$\uparrow$ & contact con.$(\%)$$\uparrow$ & diversity$\uparrow$\\ 
     \midrule
     position & 0.330 & 70.0 & 84.4  & 2.41 \\
     velocity & 0.188 & 76.1 & 89.3  & 1.95 \\
     \hline
     Ours  & 0.263 & 73.3 & 86.6 & 2.55 \\
    \bottomrule
    \end{tabular}
    }
    }
    \caption{\textbf{Flexible transfer features.} Our method support diverse motion matching features.}
    \label{tab:flexible_feature}
\end{table}

\subsection{Flexible \camera{Transfer} Features}

One of the key component of our method is the motion feature matching, whose matching feature is quite flexible. As shown in Alg.~\ref{alg:transfer}, the defined motion feature of 6D rotation is the default setting. However, the 3D loco-position or joint velocity features can also be used as matching features. As shown in~\cref{tab:flexible_feature}, our method enjoys good performance with diverse matching features. Especially, the velocity feature even works better than 6D rotations and with less diversity. This is mainly because velocity is a straightforward indicator in animation, capturing motion details sensitively, which also makes the matching process more deterministic.

\subsection{Key-frame Cross-topology \camera{Transfer}}

We found that our method can also be used in key-frame interpolation when given only some animation frames from the source skeletons. In this case, the invisible frames are initialized with the noise. Although these frames are matched blindly initially, the completion of the whole sequence can be reached in $L$ turns, according to \cref{alg:transfer}. As shown in \cref{fig:key-frame}, our method robustly works even when the source animation is very sparse in the temporal axis, not only in the spatial axis. Specifically, the retargeted unseen frames are well aligned with the source motion. This capability demonstrates that our method remains effective even when the source motion is corrupted or poorly crafted.

\subsection{Ablation Study}

\noindent \textbf{\textit{How can $\alpha$, patch size $Ps$, and $L$ affect the quality and diversity?}}
We conducted ablations over the patch‐matching hyperparameters to evaluate their impact on reconstruction quality and output diversity. Here, we do not distinguish two types of skeletal differences. The blending weight $\alpha$ controls the strength of adherence to the bound joints’ motion. When $\alpha$ is set too small (\textit{e.g.}\ 0.3), retargeted motion becomes erratic and contact consistency drops. When set too large (\textit{e.g.}\ 0.95), motion becomes overly ``locked'' to the source, reducing diversity significantly. 
Varying the patch size $P_S$ shows that increasing $P_S$ results in comparable quality, and less diversity. This indicates that a too large horizon may include more complex motion patterns for matching. The decrease in $P_S$ compromises the consistency and quality, because the short temporal window size cannot capture the semantically fruitful motion feature. 
Increasing $L$ (from 1 to 5 iterations) improves temporal consistency and motion and motion quality. However, excessively large $L$ offers minimal further improvement and incurs greater computational cost. We therefore adopt $L = 3$ as a reliable default across both similar‐skeleton and cross‐species settings.

\begin{figure*}[!t]
    \centering
    \includegraphics[width=\linewidth]{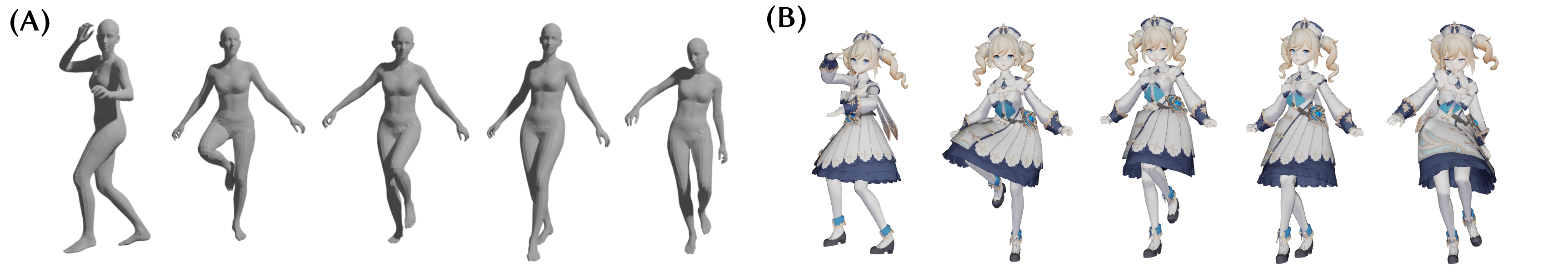}
    \caption{Key frames of the source captured SMPL-based motion (A) and the retargeted motion (B) on a character.}
    \label{fig:key-frame-smpl-any}
\end{figure*}

\begin{table}[!t]
    \centering
    \resizebox{\linewidth}{!}{
    \begin{tabular}{c|ccccc}
    \toprule
     Settings & FID$\downarrow$ & freq. align$(\%)$$\uparrow$ & contact con.$(\%)$$\uparrow$ & diversity$\uparrow$\\ 
    \midrule
     Ours ($L=1$) & 0.412 & 65.4 & 70.4 & 2.64 \\
     Ours ($L=5$) & 0.255 & 73.8 & 87.0 & 2.50 \\
     \hline
     Ours ($Ps=3$) & 0.309 & 71.1 & 83.3 & 2.60 \\
     Ours ($Ps=17$) & 0.264 & 70.0 & 80.5 & 1.95 \\
     \hline
     Ours ($\alpha=0.6$) & 0.359 & 60.1 & 67.1 & 4.19 \\
     Ours ($\alpha=0.7$) & 0.344 & 65.5 & 75.4 & 3.51 \\
     Ours ($\alpha=0.8$) & 0.298 & 71.8 & 83.3 & 2.60 \\
     Ours ($\alpha=0.95$) & 0.257 & 75.5 & 89.1 & 0.12 \\
     \hline
     Ours (default) & 0.263 & 73.3 & 86.6 & 2.55 \\
    \bottomrule
    \end{tabular}
    }
    \caption{\textbf{Abalation study of different settings.} The default setting of the \methodname is $L=3$, $P_S=11$, and $\alpha=0.85$.}
    \label{tab:abaltion}
\end{table}

\begin{table}[!t]
    \centering
    \resizebox{\linewidth}{!}{
    \begin{tabular}{c|cccc}
    \toprule
         \multicolumn{5}{c}{Similar Skeleton}\\ \cline{1-5}
         &  FID$\downarrow$ & freq. align$(\%)$$\uparrow$ & contact con.$(\%)$$\uparrow$ & diversity$\uparrow$\\ 
    \midrule
         Ours (auto.) & 0.039 & 94.5 & 91.7 & {3.25} \\
         Ours (directly copy) & 0.024 & 97.1 & 95.9 & 2.99 \\ \hline
         Ours (default)& {0.033} & {96.2} & {93.5} & {3.20} \\
    \hline 
    \hline
         \multicolumn{5}{c}{Cross-species Skeleton}\\ \cline{1-5}
         &  FID$\downarrow$ & freq. align$(\%)$$\uparrow$ & contact con.$(\%)$$\uparrow$ & diversity$\uparrow$\\ 
    \midrule
         Ours (auto.) & 0.666 & 86.7 & 72.4 &  1.98\\
         Ours (directly copy) & 0.702  & 88.0 & 75.3 & 1.66\\ \hline
         Ours (default)& {0.492} & {90.3} & {79.7} & {1.90} \\
    \bottomrule
    \end{tabular}
    }
    \caption{\textbf{Ablation study of binding mechanism and transfer strategies.} We compare automatic and manual bone binding, as well as direct copying bound motion features from the source after executing \cref{alg:transfer}.}
    \label{tab:auto_and_copy}
\end{table}

\noindent \textbf{\textit{Bone binding automatically \textit{v.s.} manually?}} 
\cref{tab:auto_and_copy} compares our default (manual) binding setting with the automatic mode using fuzzy graph matching (algorithm in \cref{sec:binding}). For similar skeletons, automatic binding incurs a minor quality drop, and contact consistency lightly falls. This shows flexible binding choices. For cross‐species \camera{transfers}, the gap widens, denoting the value of expert‐verified correspondences when topologies differ greatly. Nonetheless, the automatic mode still produces reasonable results without user effort, making it a practical solution.

\noindent \textbf{\textit{Can binding bones be directly copied with source motion features?}} 
We also evaluated a na\"ive strategy that directly copies source binding joint features to bound target joints (``Ours (directly copy)'') after patch matching in \cref{alg:transfer}. As shown in Table~\ref{tab:auto_and_copy}, direct copying achieves slightly higher \camera{transfer} quality on cross-similar-skeleton \camera{transfer}. In cross‐species \camera{transfer}, the drawback is amplified. These results confirm that we can directly copy the bound joint motion to the target skeleton when two skeletons share similar skeleton topologies (\textit{e.g.}\ SMPL to other human characters).

\noindent \textbf{\textit{What is the best binding rate for bone correspondence?}} Due to page limits, we answer this question in \cref{sec:bind}.

\begin{figure}[!t]
    \centering
    \includegraphics[width=\linewidth]{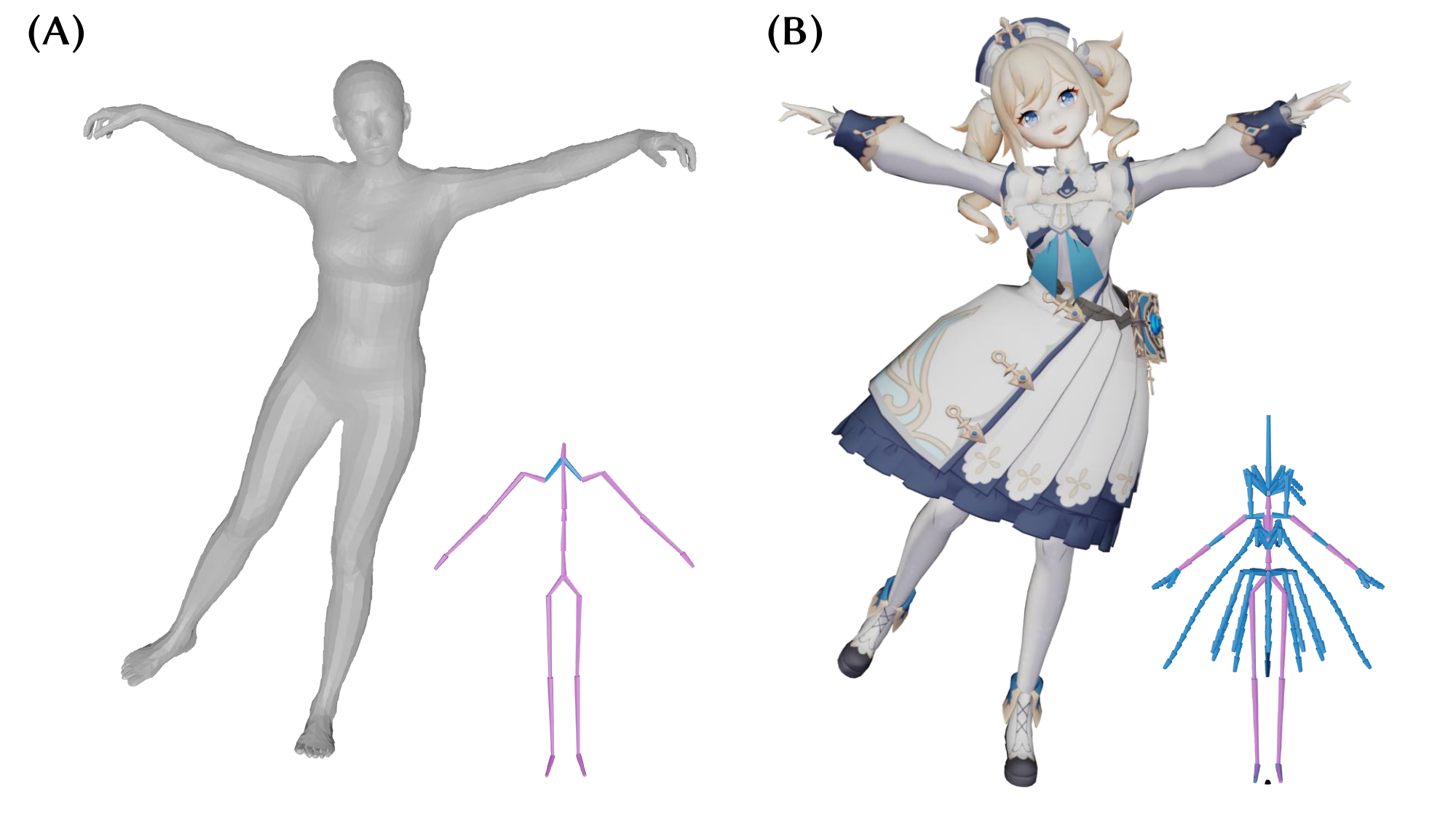}
    \caption{\textbf{Application: Lifting SMPL-based motion to any characters.} (A) The SMPL-based source motion captured from video. (B) The character with 331 joints was retargeted by the SMPL-based motion.  The bound 21 joints are in purple. The retargeted frames are shown in \cref{fig:key-frame-smpl-any}.}
    \label{fig:smpl2any}
\end{figure}
\section{Application: Lifting A SMPL-based Motion to ``Any'' Character}

A direct and realistic application of \methodname is lifting the topological skeleton structure into more complex types. For instance, the generated SMPL-based motion with text or music~\cite{motiongpt,finedance}, even captured from videos~\cite{humanmm,gvhmr}, cannot be directly retargeted to the characters used in industrial characters, whose skeletons have higher DoFs. As shown in \cref{fig:smpl2any}, we retarget a SMPL-based motion captured by HumanMM~\cite{humanmm} to a new character with 331 joints, including skirt and hair points. This shows a strong application of retarget motion to private characters from generated/captured SMPL~\cite{smpl} motions, whose algorithm is the most widely studied by the research community.  

\section{User Interface}
\label{sec:add-on}
\begin{figure}[!t]
    \centering
    \includegraphics[width=\linewidth]{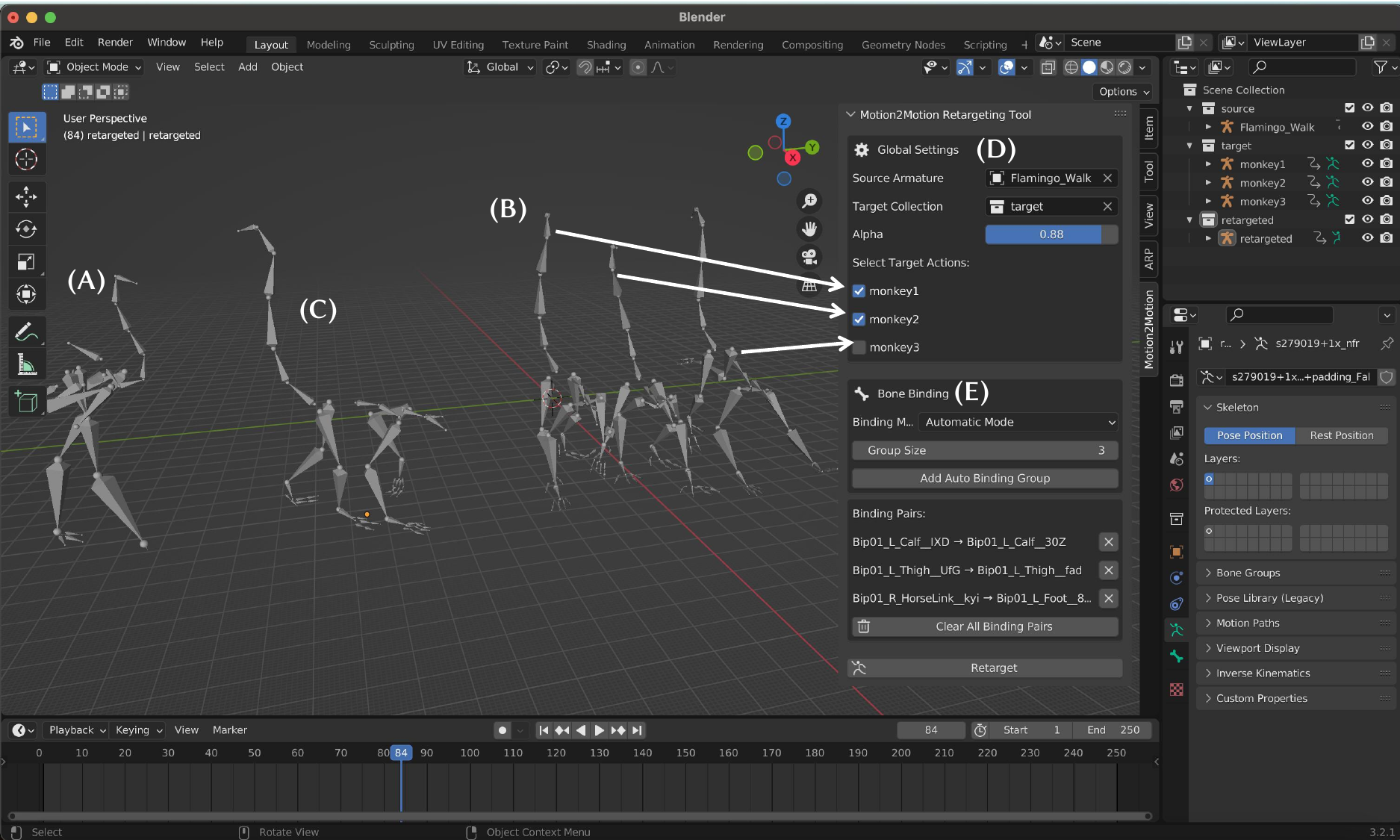}
    \caption{
        \textbf{User interface: Blender add-on for \methodname.} (A) Source motion of the flamingo. (B) Users can select few-shot samples from three motions from the target monkey skeleton. (C) The retargeted motion of the monkey. (D) Global settings of the skeleton choices. (E) The bone binding module (options: automatic or manual binding modes). 
    }
    \vspace{-1em}
    \label{fig:add-on}
\end{figure}

To demonstrate the practical viability of our \methodname, we develop a Blender add-on that integrates seamlessly with the native animation workflow, shown in \cref{fig:add-on}. The user first loads a source motion (A), for example a flamingo walking motion, then selects a few target-skeleton reference clips (B). The global panel (D) lets users specify the source armature, target motions, and blending weight $\alpha$, while the bone-binding module (E) supports both automatic binding and manual adjustment of bone pairs. After clicking the ``\camera{transfer}'' button, the retargeted motion (C) will be synthesized in real time. This intuitive interface streamlines the cross-topology motion \camera{transfer} application and indicates our method can be deployed directly within the animation creation workflow.

\vspace{-0.5em}
\section{Conclusion}

\noindent \textbf{\textit{Conclusion.}} In this paper, we propose a novel framework, \methodname, to retarget an animation from a source skeleton to a target one, requiring only very sparse bone correspondence. The proposed framework works in a training-free fashion with real-time efficiency on CPU-only devices. Our algorithm works in a patchwise motion matching mechanism, supporting flexible motion features for matching. Extensive experimental results indicate that our method is quite robust in in-species and cross-species motion \camera{transfer}, demonstrating its applicability in diverse downstream tasks, especially \camera{transfer} SMPL-based motion to ``any'' character. We also show the interaction interface of \methodname in applications of animation creation pipelines.

\noindent \textbf{\textit{Failure cases.}} Though our method works well in some scenarios, it may fail in some cases. If the target examples are quite semantically different from the source motion (kungfu \vs dancing), the retargeted result fails. Despite this, the community still lacks a reasonable solution to this. At present, requiring some key frames as target examples by humans is a relatively applicable method for this. 

\noindent \textbf{\textit{Limitations and future work.}} While this work represents a pioneering effort in cross-topology motion \camera{transfer}, it still has limitations. Our algorithm relies on one- or few-shot target motions, which imposes laborious demands on the animation creation pipeline. Although this level of data requirement is minimal by current community standards, we will continue to explore more data-efficient and lightweight approaches in future work.

\section*{Acknowledgement}

The \methodname author team would like to acknowledge all program committee members for their extensive efforts and constructive suggestions. In addition, Weiyu Li (HKUST), Shunlin Lu (CUHK-SZ), and Bohong Chen (ZJU) had discussed with the author team many times throughout the process. The author team would like to convey sincere appreciation to them as well. 

\bibliographystyle{ACM-Reference-Format}
\bibliography{src/ref}

\newpage
\appendix

\section{Joint Binding Algorithm}
\label{sec:binding}

Motivated by \citet{sun2020rapidmatch}, we design an automatic bone binding method for users. Let $|\mathcal{N}|$ be the number of nodes in the tree and $L$ be the maximum path length. For each node, the algorithm traces upward for at most $L$ steps, resulting in a time complexity of $\mathcal{O}(|\mathcal{N}|L)$. Each valid path contains at most $L$ nodes, and up to $|\mathcal{N}|$ such paths can be collected, leading to a space complexity of $\mathcal{O}(|\mathcal{N}|L)$. The algorithm is detailed in \cref{alg:trace_upward}.

Before excusing the matching algorithm above, we calculate the unit direction vector of each bone for the source and target skeletons. For the source and target skeletons, given the same path length $L$, we construct two node sets for each ($\mathcal{C}_s$ and $\mathcal{C}_t$). We calculate the cosine similarity of two direction vectors from two sets pairwisely. Accordingly, we calculate all L-length subgraph similarities by the proposed similarity measurement. We return the pair with the highest similarity to the user as the binding bones, and feed them into the \methodname.

\begin{algorithm}[!h]
\SetAlgoLined
\caption{Trace $L$-length Paths of a skeleton tree}
\label{alg:trace_upward}
\KwIn{Skeleton tree $\mathbb{T}$ with a node set $\mathcal{N}$, path length $L$.}
\KwOut{A set of valid chains $\mathcal{C}$ with length $L$.}

\SetKwFunction{FTraceUp}{TraceUp}
\SetKwProg{Fn}{Function}{:}{}
\Fn{\FTraceUp{$\texttt{node}$}}{
    $\texttt{path} \leftarrow \{\texttt{node}\}$\\
    $\texttt{current} \leftarrow \texttt{node}$\\

    \While{$|\texttt{path}| < L$ \textbf{\text{and}} \texttt{has\_parent(current)}}{
        $\texttt{parent} \leftarrow \texttt{get\_parent(current)}$\\
        $\texttt{path.append(parent)}$\\
        $\texttt{current} \leftarrow \texttt{parent}$\\
    }
    
    \If{$|\texttt{path}| = L$}{
        \texttt{return path} \tcp*{find a $\texttt{path}$}
    }
}
\Return{$\{ \ \}$} \\
\BlankLine
$\mathcal{C} \leftarrow \{ \ \}$ \tcp*{initialize $\mathcal{C}$}
\ForEach{$\texttt{node} \in \mathcal{N}$}{
    $\texttt{path = TraceUp(node)}$\\
    \If{$|\texttt{path}| > 0$ }{
        $\mathcal{C}.\texttt{append(path)}$ \tcp*{append the $\texttt{path}$}
    }
}
\Return{$\mathcal{C}$}
\end{algorithm}

\newpage

\section{Desired Binding Rate} 
\label{sec:bind}

\begin{figure}[h]
    \centering
    \begin{subfigure}{\linewidth}
        \includegraphics[width=\linewidth]{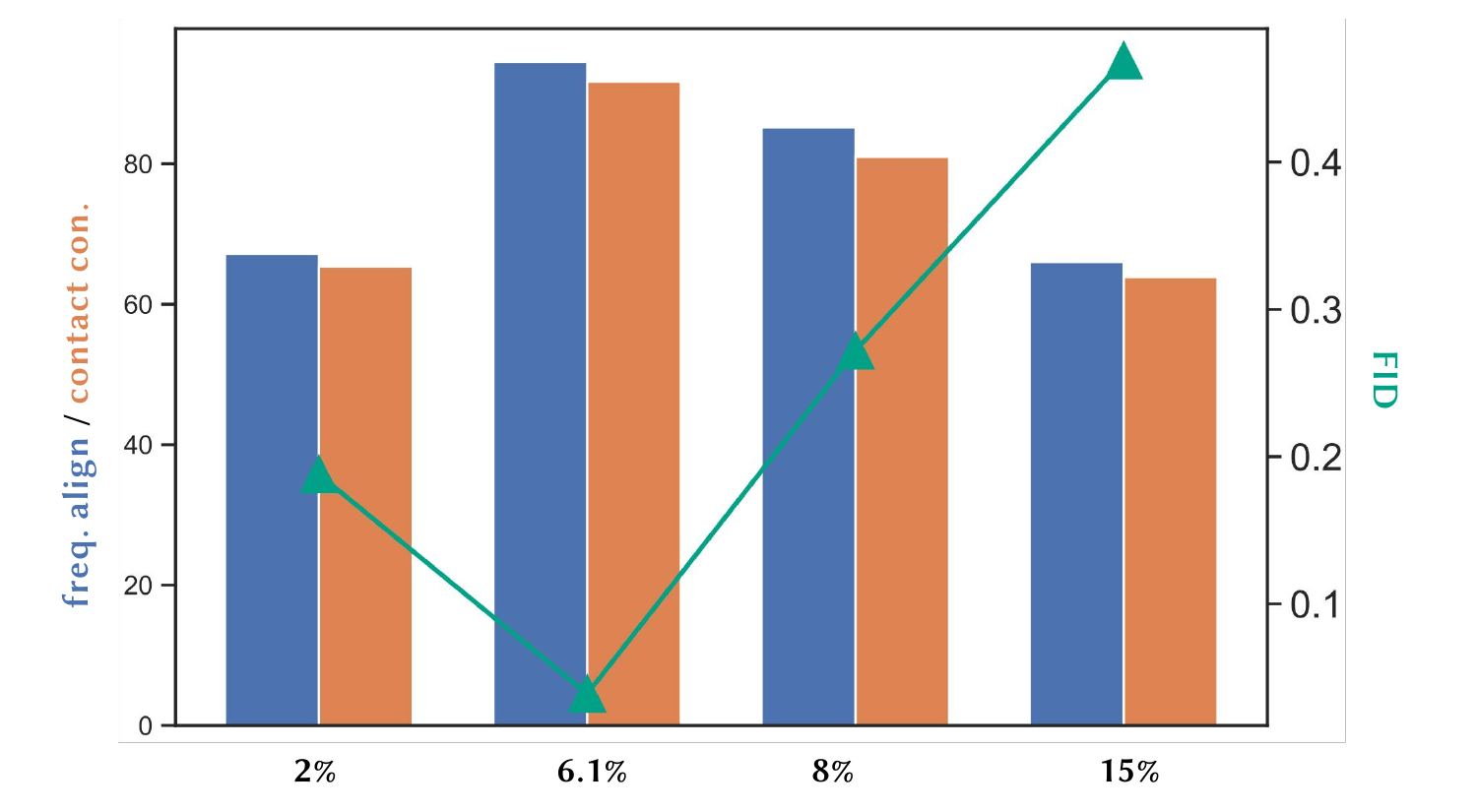}
        \caption{(A) Similar skeleton motion \camera{transfer}.}
        \label{fig:ins}
    \end{subfigure}
    \begin{subfigure}{\linewidth}
        \includegraphics[width=\linewidth]{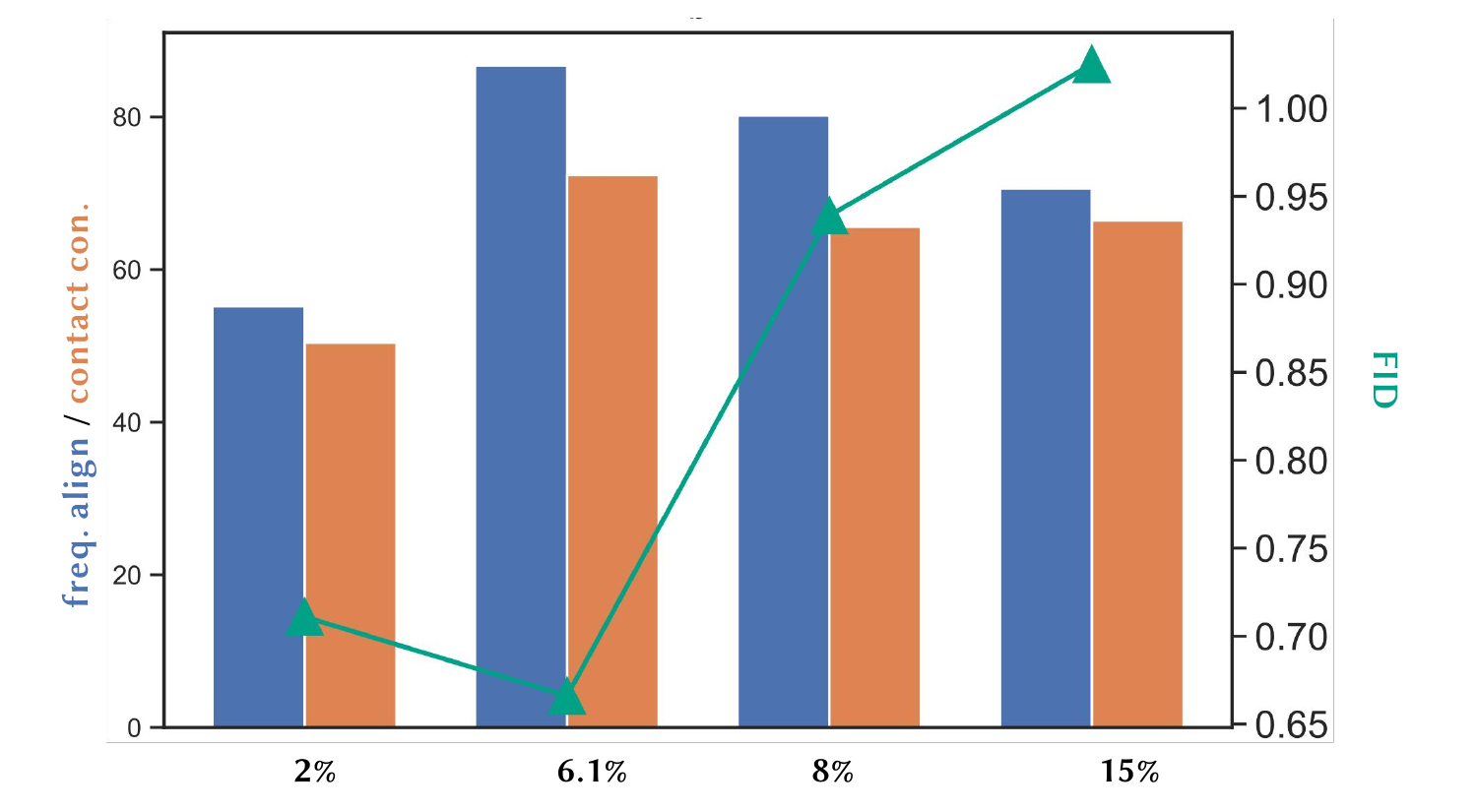}
        \caption{(B) Cross-species skeleton motion \camera{transfer}.}
        \label{fig:outs}
    \end{subfigure}
    \caption{Comparison of different binding rates.}
    \label{fig:ins_outs}
\end{figure}

Here, we raise a research question: \textit{What is the best binding rate for bone correspondence?} To answer this question, we ablate a group of binding joint numbers. To simplify and align the setting, we set the binding mode to automatic. We calculate the bonding rate from 2\% to 15\%, and our default setting in the automatic binding mode is 6.1\%. 
As can be seen in \cref{fig:ins_outs}, different binding rates have varying effects on both similar skeletons and cross-species motion \camera{transfer}. Our default setting of 6.1\% achieves a favorable balance between motion fidelity and anatomical plausibility. When the binding rate is too small, it is hard for the algorithm to find the coherence between two skeletons. If the rate is too large, it will introduce some mismatched correspondences, resulting in some inaccuracy. That is to say, sometimes ``\textit{less is more}''. \textbf{Fortunately}, our binding algorithm is interactive to choose the bindings, which is adjustable for users. 

\newpage

\section{User Study Template}

\begin{figure}[!h]
    \centering
    \includegraphics[width=\linewidth]{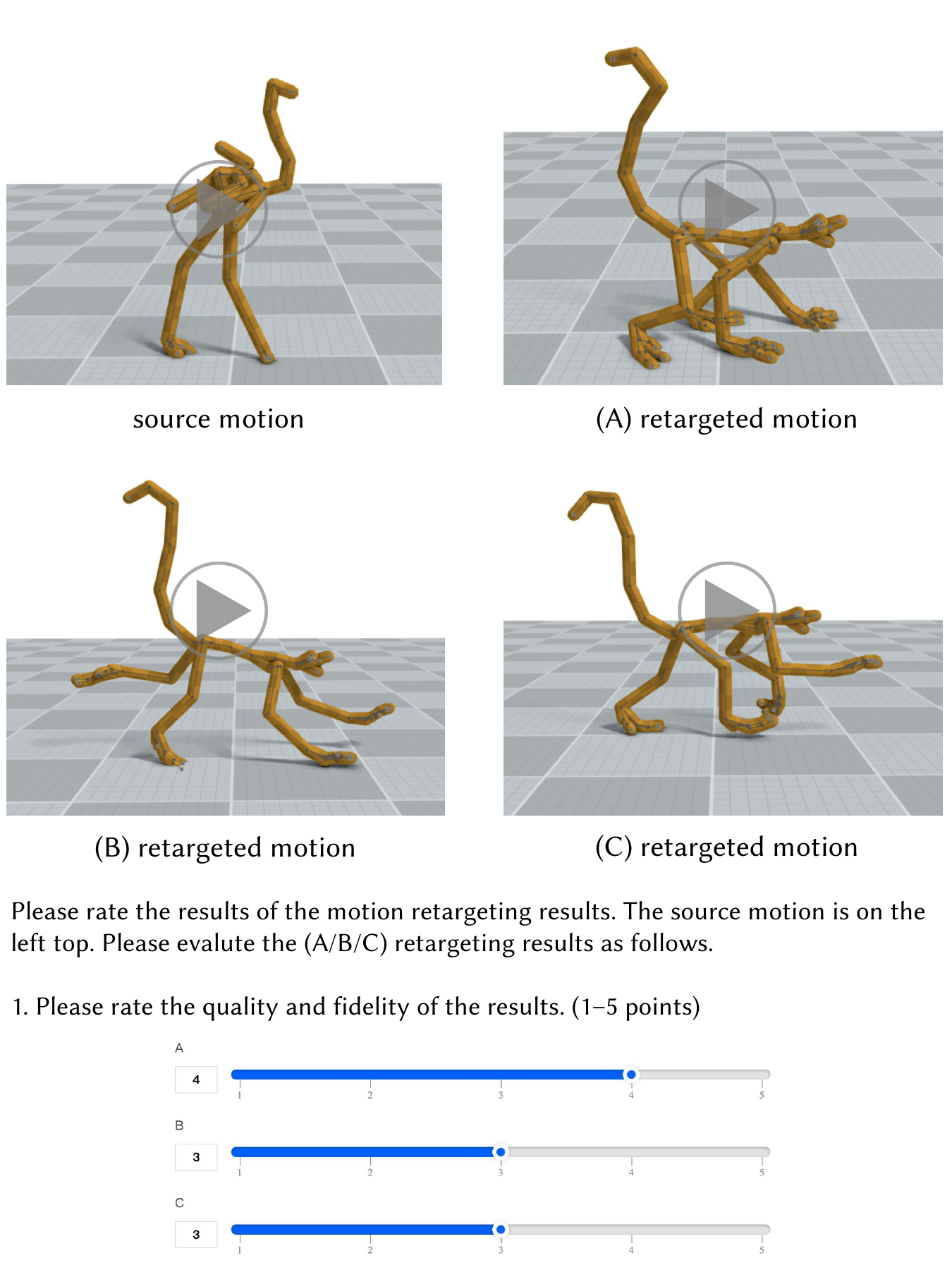}
    \caption{User study template. }
    \label{fig:user_study}
\end{figure}

We provide the template of the user study cases in \cref{fig:user_study}. The participants are asked to watch the source motion (top-left) and evaluate three retargeted results (A/B/C) based on motion quality and coherence. Each result is scored on a 5-point Likert scale, where 5 indicates the highest quality. This setup allows for a consistent and fair comparison of different \camera{transfer} methods.

\clearpage
\end{document}